%% file: ms.tex
\documentclass[11pt]{gsasthesis}

\usepackage{etex}
\usepackage{todonotes}
\usepackage[margin={1.2in}]{geometry}
\usepackage[titletoc]{appendix}
\usepackage{rotating}
\usepackage{longtable}
\usepackage{natbib,bibentry}
\nobibliography*
\usepackage[sc]{mathpazo}
\usepackage{courier}
\usepackage[utf8]{inputenc}
\usepackage[T1]{fontenc}
\usepackage[english]{babel}
\usepackage{blindtext}
\usepackage{amsmath}
\usepackage{microtype}
\usepackage[stable,multiple]{footmisc}
\usepackage{csquotes}
\usepackage{hyperref,url,booktabs,amsfonts,nicefrac,microtype,amsmath,amssymb,mathtools,xcolor,graphicx,float,fancyhdr}
\usepackage{multirow}
\usepackage{tabu}
\usepackage{makecell}
\usepackage{arydshln}
\usepackage{hhline}
\usepackage[export]{adjustbox}

\newcommand{\norm}[1]{\left\lVert#1\right\rVert}
\newcommand{\inR}[2]{\in \mathbb{R}^{#1 \times #2}}
\DeclareMathOperator*{\argmin}{arg\,min}

\RequirePackage[font=small,format=plain,labelfont=bf,textfont=it]{caption}
\title{Training Machine Learning Models by Regularizing their Explanations}
\author{Andrew Slavin Ross}
\degreename{Master of Engineering}
\degreefield{Computational Science and Engineering}
\department{The Institute for Applied Computational Science}
\degreemonth{May}
\degreeyear{2018}
\principaladvisor{Finale Doshi-Velez}

\begin{document}

\pagenumbering{roman}

\thesistitlepage
\copyrightpage
\begin{abstract}
  Neural networks are among the most accurate supervised learning
methods in use today.  However, their opacity makes them difficult to
trust in critical applications, especially when conditions in training
may differ from those in practice.
   Recent efforts to develop \textit{explanations} for neural networks and machine learning models more generally have produced tools to
  shed light on the implicit rules behind predictions. These tools can help us identify when
  models are right for the wrong reasons. However, they do not always
  scale to explaining predictions for entire datasets, are not always at the right
  level of abstraction, and most importantly cannot correct the
  problems they reveal.  In this thesis, we explore the possibility of training
  machine learning models (with a particular focus on neural networks) using explanations themselves. We consider
  approaches where models are penalized not only for making incorrect predictions but also for providing explanations that are either inconsistent with
  domain knowledge or overly complex.  These methods let us train models
  which can not only provide more interpretable rationales for their
  predictions but also generalize better when training data is confounded or
  meaningfully different from test data (even adversarially so).
\end{abstract}

\renewcommand{\contentsname}{\protect\centering\protect\Large Contents}
\renewcommand{\listtablename}{\protect\centering\protect\Large List of Tables}
\renewcommand{\listfigurename}{\protect\centering\protect\Large List of Figures}
\tableofcontents
\pagenumbering{arabic}

\addcontentsline{toc}{chapter}{Introduction}
\chapter{Introduction}\label{ch:intro}
\input{intro}

\chapter{Right for the Right Reasons\footnotemark}\label{ch:1}
\footnotetext{Significant portions of this chapter also appear in \bibentry{rrr}.}
\input{chapter1}

\chapter{Interpretability and Robustness\footnotemark}\label{ch:2}
\footnotetext{Significant portions of this chapter also appear in \bibentry{adv}.}
\input{chapter2}

\chapter{General Explanation Regularization}\label{ch:3}

In the previous two chapters, we introduced the idea of explanation regularization,
and showed that we could use this to obtain models that were both simpler and more robust to
differences between training and test conditions. However, we obtained all of those results
just with L2 input gradient penalties. Although gradients have special
importance in differentiable models such as neural networks,
they have major limitations, especially when the kind of constraints we would
like to impose on an explanation are abstract in a way we cannot easily
relate back to input features. So in this chapter, we outline
promising avenues towards more general forms of explanation regularization.

\section{Alternative Input Gradient Penalties}

Before we leave input gradients behind altogether, it is worth considering what else we can
do with them besides simple L2 regularization.

\subsection{L1 Regularization}

In Chapter \ref{ch:2}, we saw that penalizing the L2 norm of our model's input
gradients encouraged gradient interpretability and prediction robustness to
adversarial examples, and drew an analogy to Ridge regression. One natural
question to ask is how penalizing the L1 norm instead would compare, which we
could understand as a form of local linear LASSO.

For a discussion of this question with application to sepsis treatment, we
refer the reader to \cite{neural-lasso}, which includes a case-study showing
how L1 gradient regularization can help us obtain mortality risk models that
are \textit{locally} sparse and more consistent with clinical knowledge.

\begin{figure}
  \begin{center}
\includegraphics[width=0.45\textwidth]{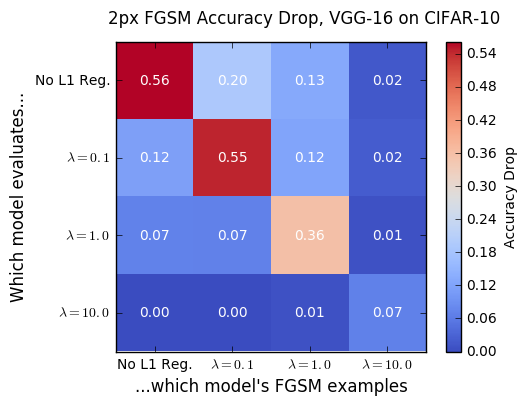}
\includegraphics[width=0.54\textwidth]{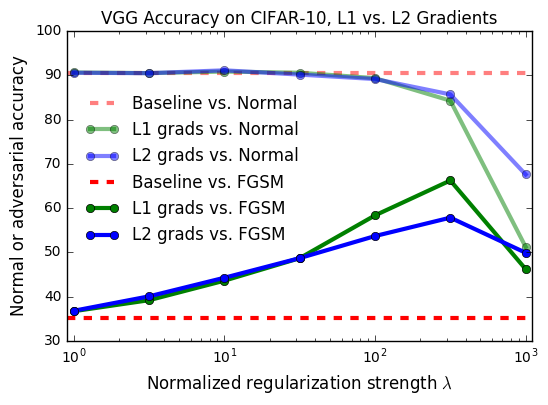}
  \end{center}
\caption{Left: Accuracy loss on CIFAR-10 FGSM examples ($\epsilon=2$px) for VGG models trained with varying levels of L1 gradient regularization. Diagonals measure white-box vulnerability and off-diagonals measure transferability.
  Right: L1 vs. L2 gradient regularization on VGG as a defense against white-box FGSM examples, 2px perturbation. The value of $\lambda$ is multiplied by 100 for the L1 regularized network to equalize penalty magnitudes (since we do not take the square root of the L2 penalty). Compared to L2, L1 gradient regularized models tend to be more robust to $l_\infty$ attacks like the FGSM, and their adversarial examples tend to be less transferable.}
 \label{fig:vgg-fgsm-accuracy}
\end{figure}

On image datasets (where input features are not individually meaningful), we do
find that L1 gradient regularization is effective in defending against
adversarial examples, perhaps more so than L2 regularization. To that end, in Figure
\ref{fig:vgg-fgsm-accuracy} we present
results for VGG-16 models on CIFAR-10, which bode favorably for L1
regularization against both white- and black-box attacks. However, although the
gradients of these models change qualitatively compared to normal models, they
are not significantly sparser than gradients of models trained with L2 gradient
regularization. These results suggest that sparsity with respect to input
features may not be a fully achievable or desirable objective for complex image
classification tasks.

\subsection{Higher-Order Derivatives}

\cite{bishop1993curvature} introduced the idea of limiting the curvature of the
function learned by a neural network by imposing an L2 penalty on the
network's second input derivatives. They note, however, that evaluating
these second derivatives increases the computational
complexity of training by a factor of $D$, the number
of input dimensions. This scaling behavior poses major
practical problems for datasets like ImageNet, whose inputs are
over 150,000-dimensional. \cite{rifai2011higher} develop a scalable workaround by
estimating the Frobenius norm of the input Hessian as $\frac{1}{\sigma^2} \mathbb{E}\left[\left|\left|\nabla_X f(x) - \nabla_X
f(x+\epsilon)\right|\right|^2_2\right]$ for $\epsilon \stackrel{iid}{\sim} \mathcal{N}(0,\sigma^2)$, which converges to the true value as $\sigma \to 0$. They then train autoencoders
whose exact gradient and approximate Hessian norms are both L2-penalized, and find that the
unsupervised representations they learn are more useful for downstream
classification tasks.
\cite{czarnecki2017sobolev} also regularize using estimates of higher-order
derivatives.

\begin{figure}
\begin{center}
\includegraphics[width=0.45\textwidth]{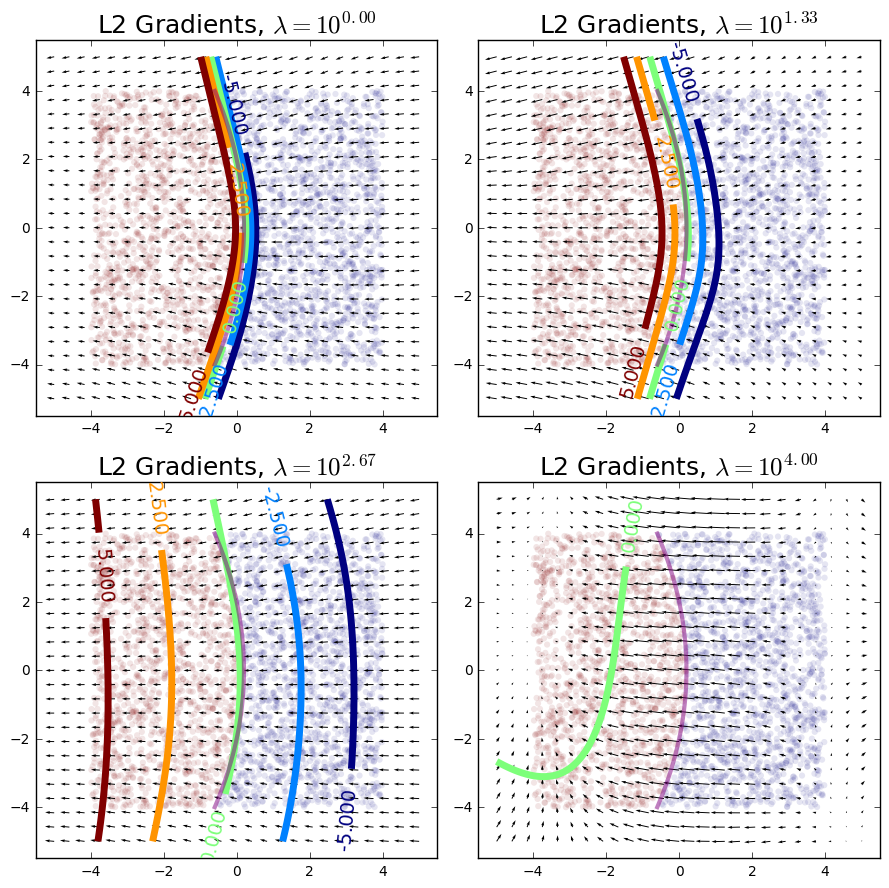}
\hspace{0.5cm}
\includegraphics[width=0.45\textwidth]{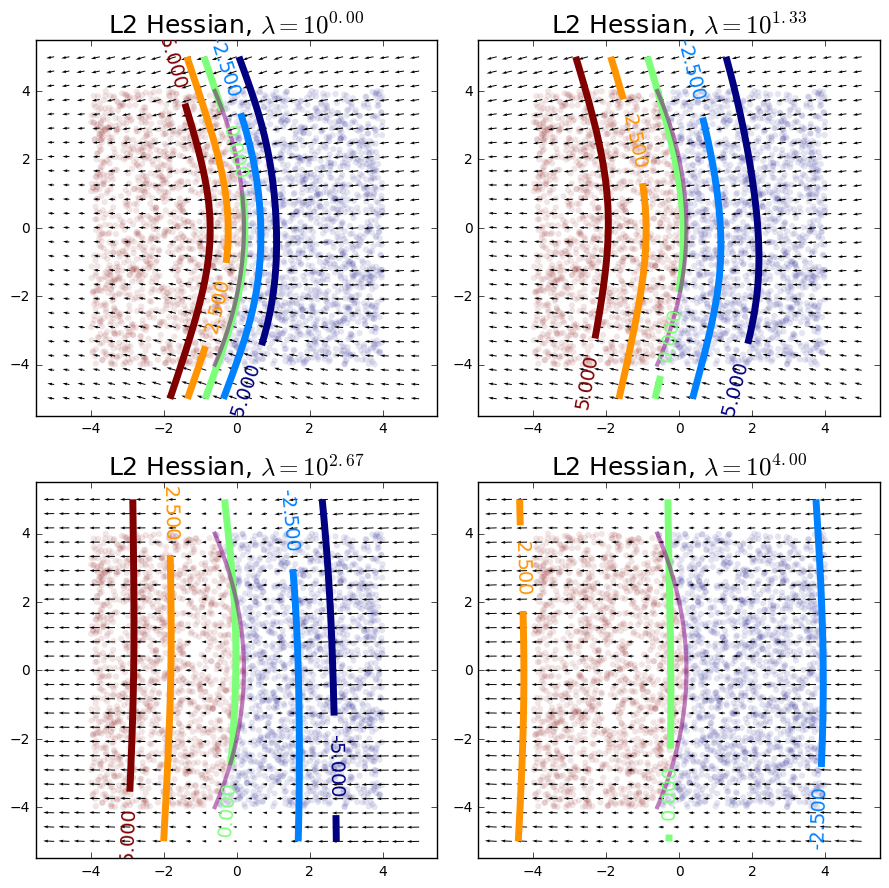}
\end{center}
\caption{Gradient regularization (left) vs. Hessian regularization (right).
Purple line indicates the true decision boundary; other lines indicate level sets
of a 10-hidden unit MLP's predicted log-odds from -5 to 5 by increments of 2.5,
with the model's decision boundary in green.
Hessian regularization can make decision boundaries wider and flatter without
triggering pathological cases.}
\label{fig:hess-reg-2d}
\end{figure}

Hessian regularization may be desirable for adversarial robustness and
interpretability as well. The results in Figure \ref{fig:hess-reg-2d} suggest that exact
Hessian regularization for an MLP on a simple 2D problem encourages the model
to learn flatter and wider decision boundaries than gradient regularization,
which could be useful for interpretability and robustness. Hessian regularization
also appears to behave more sensically even when the penalty term is much larger than
the cross entropy. By contrast, in this regime, gradient regularization starts
pathologically seeking areas of the input space (usually near the edges of the training distribution)
where it can set gradients to 0.

\section{Heftier Surrogates}

While input gradient-based methods are appealing because of their close
relationship to the shape and curvature of differentiable models' decision surfaces, they are
limited by their locality and humans' inability to express abstract desiderata
in terms of input features. This second limitation in particular prevents us from optimizing
for the kind of simplicity or diversity humans find intuitive.
Therefore, in the next sections we explore
ways of training models using more complex forms of explanation.

One common way of explaining complicated models like neural networks is by
distilling them into surrogate models; decision trees are a particularly
popular choice \cite{craven1996extracting}. However, these decision trees must
sometimes be quite deep in order to accurately explain the associated
networks, which defeats the purpose of making predictions interpretable.
To address this problem, \cite{wu2017beyond} optimize the underlying \textit{neural networks}
to be accurately approximatable by \textit{shallow} decision trees. Performing
such an optimization is difficult because the process of distilling a network
into a decision tree cannot be expressed analytically, much less
differentiated. However, they approximate it by
training a \textit{second} neural network to predict the depth of the decision
tree that would result from the first neural network's parameters. They then
use this learned function as a differentiable surrogate of the true
approximating decision tree depth.  Crucially, they find a depth regime where
their networks can outperform decision trees while remaining explainable by
them. Although they only try to minimize the approximating decision tree depth,
in principle one could train the second network to estimate other characteristics
of the decision tree related to simplicity or consistency with domain knowledge
(and optimize the main network accordingly).

\section{Examples and Exemplars}

Another popular way of explaining predictions is
with inputs themselves.  k-Nearest Neighbors (kNN) algorithms are easy to
understand since one can simply present the neighbors, and techniques have
recently been proposed to perform kNN using distance metrics derived from
pretrained neural networks \citep{papernot2018deep}.  More general methods
involve sparse graph flows between labeled and unlabeled inputs
\citep{rustamov2017interpretable} or optimization to find small sets of
prototypical inputs that can be used for cluster characterization or
classification \citep{kim2014bayesian}, even within neural networks
\citep{li2017deep}.  There has also been recent work on determining which
points would most affect a prediction if removed from the training set
\citep{koh2017understanding}. These approaches have both advantages and disadvantages.
Justifying predictions based on input similarity
and difference can seem quite natural, though it can also be confusing or
misleading when the metric used to quantify distance between points does not
correspond to human intuition. Influence functions shed light on model sensitivities
that are otherwise very hard to detect, but they are also very
sensitive to outliers, leading to sometimes inscrutable explanations.

However, it seems straightforward at least in principle to implement
example-based explanation regularization.  For example, we could train neural
networks with annotations indicating that certain pairs of examples should be
similar or dissimilar, and penalize the model when their intermediate
representations are relatively distant or close (which might require altering
minibatch sampling to keep paired examples together if
annotations are sparse).  Although influence functions may be too
computationally expensive to incorporate into the loss functions of large
networks, it seems useful in principle to specify that certain examples
should be particularly representative or influential in deciding how to classify others.

\section{Emergent Abstractions}

Stepping back, the level of abstraction at which we communicate the reason
behind a decision significantly affects its utility, as \cite{explanations}
notes:

\begin{displayquote}
Explanations... suffer if presented at the wrong level of detail. Thus, if
asked why John got on the train from New Haven to New York, a good explanation
might be that he had tickets for a Broadway show. An accurate but poor
explanation at too low a level might say that he got on the train because he
moved his right foot from the platform to the train and then followed with his
left foot. An accurate but poor explanation at too high a level might say that
he got on the train because he believed that the train would take him to New
York from New Haven.
\end{displayquote}

The explanations we have considered so far have been in terms of input
features, entire inputs, or simple surrogates. However, sometimes humans
seek to know the reasons behind predictions at levels of abstraction these
forms cannot capture. If we really want to create interpretable interfaces for
training and explaining machine learning models, humans and models will need to
speak a common language that permits abstraction.

This may seem like a daunting task, but there has been important recent
progress in interpreting neural networks in terms of abstractions that emerge
during training.  \cite{bau2017network} introduce a densely labeled image
dataset. They train convolutional neural networks on a top-level classification
task, but also include lower-level sublabels that indicate other features in
the image. They measure the extent to which different intermediate nodes in
their top-level label classifiers serve as exclusive ``detectors'' for
particular sublabels, and compare the extent to which different networks learn
different numbers of exclusive detectors. They also categorize their sublabels
and look at differences in which \textit{kinds} of sublabels each network
learns to detect (and when these detectors emerge during training).

\cite{kim2017tcav} provide a method of testing networks' sensitivity to
concepts as defined by user-provided sets of examples.
Concretely, they train a simple linear classifer at each layer to
distinguish between examples in the concept set and a negative set. They
reinterpret the weights of this linear classifier as a ``concept activation
vector,'' and take directional derivatives of the class logits with respect to
these concept activations. Repeated across the full dataset
for many different concepts, this procedure outputs a set of concept
sensitivity weights for each prediction, which can be used for explanation or
even image retrieval.

The previous two methods require manual human selection of images corresponding
to concepts, and they do not guarantee meaningful correspondence between these
concepts and what the network has learned. Feature visualization
\citep{olah2017feature} takes a different approach and attempts to understand
what the network has learned on its own terms. In particular, it tries to
explain what (groups of) neuron(s) learn by optimizing images to maximize (or
minimize) their activations. It can also optimize sets of images to jointly
maximize activations while encouraging diversity. This process can be useful
for obtaining an intuitive sense of (some of) what the model has learned, especially if
the neurons being explained are class logits. However, it also leads to an
information overload, since modern networks contain millions of neurons and an
effectively infinite number of ways to group them.  To that end,
\cite{olah2018the} use non-negative matrix factorization (NMF) to learn a small
number of groups of neurons whose feature visualizations best summarize the
entire set. Feature visualizations of neuron groups obtained by NMF tend to
correspond more cleanly to human-interpretable concepts, though again there is no
guarantee this will occur. \cite{olah2018the} also suggest that incorporating
human feedback into this process could lead to a method to train models to make
decisions ``for the right reasons.''

The above cases either take humans concepts and try to map them to network
representations or take network ``concepts'' and try to visualize them so
humans can map them to their own concepts. But they do not actually try to
\textit{align} network representations with human concepts. However,
there has been significant recent interest in training models to learn
\textit{disentangled representations}
\citep{chen2016infogan,higgins2016beta,siddharth2017learning}. Disentangled
representations are often described as separating out latent factors that
concisely characterize important aspects of the inputs but which cannot be
easily expressed in terms of their component features. Generally, disentangled
representations tend to be much easier to relate to human-intuitive concepts
than what models learn when only trained to minimize reconstruction or prediction
error.

\begin{figure}
  \begin{center}
    \includegraphics[width=0.49\textwidth]{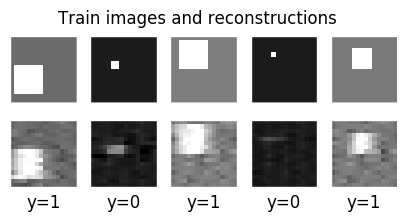}
    \includegraphics[width=0.49\textwidth]{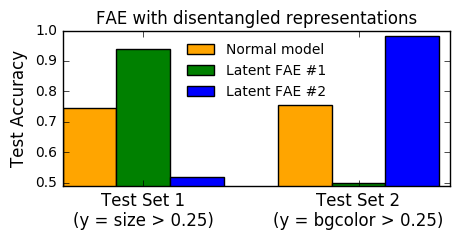}
\end{center}
\caption{Accuracies of a normal model and two models trained using find-another-explanation in a disentangled latent space (right) a toy image dataset that confounds background color and square size in training (left) but decouples them in test. Performing find-another-explanation in a latent space allows us to learn models that make predictions for conceptually different reasons, which is reflected in their complementary accuracies on each version of the test set.}
  \label{fig:latent-fae}
\end{figure}

These advances in bridging human and neural representations could have major
payoffs in terms of interpreting models or optimizing them to make predictions
for specific reasons. Suppose we are interested in testing a classifier's
sensitivity to an abstract concept entangled with our input data. If we
have an autoencoder whose representation of the input disentangles the
concept into a small set of latent factors, then for a specific input, we
can encode it, decode it, and pass the decoded input through the classifier,
taking the gradient of the network's output with respect to the latent factors
associated with the concept. If we fix the autoencoder weights but not the
classifier weights, we can use this differentiable concept sensitivity score
to apply our ``right for the right reasons'' technique from Chapter \ref{ch:1}
to encourage the classifier to be sensitive or insensitive to the \textit{concept}.

We present a preliminary proof of concept of this idea in Figure
\ref{fig:latent-fae}.  In this experiment, we construct a toy dataset of
images of white squares with four true latent factors of variation: the size of
the square, its x and y position, and the background color of the image.
In training, background color and square size are confounded; images either have
dark backgrounds and small squares or light backgrounds and large squares (and either one
can be used to predict the label). However, we create two versions of the test set where
these latent factors are decoupled (and only one predicts the label). This is analogous to the parable in our introduction with squares representing tanks and background colors representing light. When we train a one-hidden layer MLP normally, it learns to implicitly use
both factors, and obtains suboptimal accuracies of about 75\% on each test set.
To circumvent this issue,
we first train a convolutional autoencoder that disentangles square size from background color
(which we do with supervision here, but in principle this can be unsupervised)
and then prepend the autoencoder to our MLP with fixed weights.
We then simultaneously train two instantiations of this network
with the find-another-explanation penalty we introduced in Section \ref{sec:fae2}.
These two networks learn to perform nearly perfectly on one test set and do no
better than random guessing on the other, which suggests they are making predictions
for different conceptual reasons. Obtaining these networks would have been very difficult
using only gradient penalties in the input space.

\section{Interpretability Interfaces}

\citet{olah2018the} describe a space of ``interpretability interfaces'' and
introduce a formal grammar for expressing explanations of neural networks (and
a systematic way of exploring designs).  They visualize this design space in a
grid of relationships between different ``substrates'' of the design, which
include groups of neurons, dataset examples, and model parameters -- the latter
of which presents an opportunity ``to consider interfaces for \emph{taking
action} in neural networks.'' If human-defined concepts, disentangled
representations, or other forms of explanation are included as additional
substrates, one can start to imagine a very general framework for expressing
priors or constraints on relationships between them. These would be equivalent
to optimizing models to make predictions for specific reasons.

\begin{figure}[ht]
  \includegraphics[width=\textwidth]{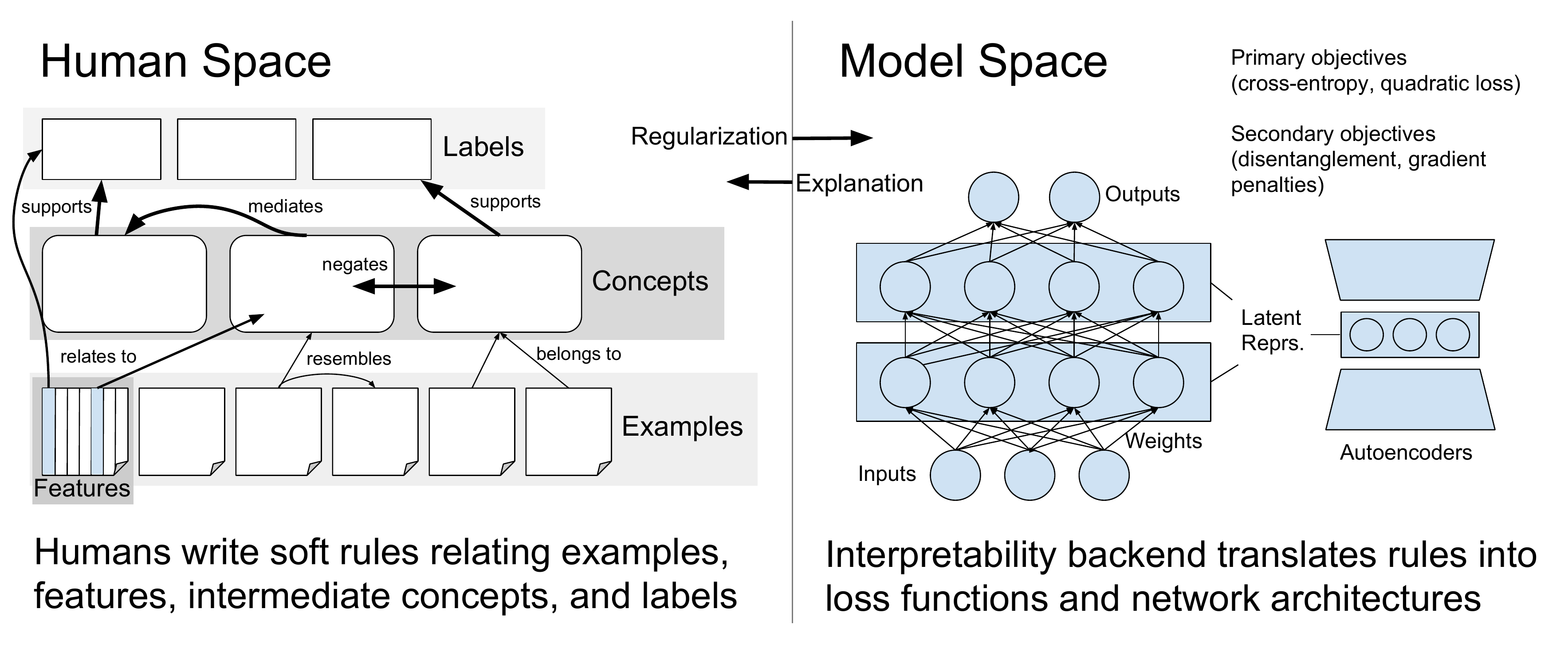}
  \caption{Schematic diagram of an interpretability interface.}
  \label{fig:interp-interf}
\end{figure}

How would humans actually express these kinds of objectives? One interface
worth emulating could be that introduced by recent but popular libraries for
weak supervision \citep{ratner2017snorkel} or probabilistic soft logic
\citep{bach2017hinge}, which is related to the well-studied topic of fuzzy
logic, a method noted for its compatibility with human reasoning
\citep{zadeh1997toward}. In these frameworks, users can specify ``soft'' logical rules
for labeling datasets or constraining relationships between atoms (or
substrates) of a system. Though users can sometimes specify that certain
rules are inviolable or highly-weighted, in general these systems assume
that rules are not always correct and attempt to infer weights for each.
While these inference problems are nontrivial, and in general there may be complex, structured
interactions between rules that are difficult to capture, the interface it exposes to users is expressive and potentially worth emulating in an interpretability interface.
For example, we could imagine writing soft rules relating:
\begin{itemize}
  \item dataset examples to each other (e.g. these examples should be conceptually similar with respect to a task)
  \item dataset examples to concepts (e.g. these are examples of a concept)
  \item features to concepts (e.g. this set of features is related to this concept, this other set is not; in this specific case, these features contribute positively)
  \item concepts to predictions (e.g. the presence of this concept makes this prediction more or less likely, except when this other concept is present)
\end{itemize}
These rules could be ``compiled'' into additional energy terms in the model's
loss function, possibly with thresholding if we expect them to be incorrect
some percentage of the time (though rules defined for specific examples may be
more reliable).
We present a schematic diagram of how a system like
this might work in Figure \ref{fig:interp-interf}.

Such a system would strongly depend on being able to define rules in terms of
abstract concepts, but such rules might not be enforcible until
the model has a differentiable, stable representations of them. However, one
could imagine pre-learning static, disentangled concept representations
that could be related back to input features. If 1:1 mappings between human concepts
and latent representations do not emerge naturally, even allowing for hierarchical
relationships \citep{hierarchicaldis}, steps could be taken to optimize model representations
to better match human understanding (e.g. using partial supervision) or to help humans
better understand model representations (e.g. using feature visualization).
This process of reaching user-model intersubjectivity might require multiple
stages of identification and refinement, but seems possible in principle.
And perhaps arriving at a shared conceptual framework for understanding
a problem is where the work of teaching and learning \textit{ought} to lie,
regardless of whether the teachers and learners are human.

\section{Discussion}

In this chapter, we discussed a number of strategies for explanation
regularization beyond the methods we used in the previous chapters. We
described simple extensions of gradient-based methods (imposing L1 and Hessian
penalties), strategies in terms of interpretable surrogates (regularizing
distilled decision trees, nearest neighbors, and exemplars), and strategies in
terms of concepts (concept activation vectors, disentangled representations,
and feature visualization). We then combined many of these strategies into a
design for an ``interpretability interface'' that could be used to
simultaneously improve neural network interpretability and incorporate domain
knowledge.

One limitation of this discussion is that we only considered classification
models and traditional ways of explaining their predictions. However, there is
a much larger literature on alternative forms of explanation and prediction
like intuitive theories \citep{gerstenberg2017intuitive} or causal inference
\citep{pearl2010causal} that is highly relevant, especially if we want to apply
these techniques to problems like sequential decisionmaking. We started this
thesis by making a point that was ``easiest to express with a story;'' even
with arbitrarily human-friendly compositional abstraction
\citep{schulz2017compositional}, flat sets of concepts may never be sufficient
in cases where users think in terms of narratives \citep{abell2004narrative}.

However, despite these limitations,
we think the works we have outlined in this chapter have started to map a
rich design space for interpreting and training machine learning models with
more than just $x$es and $y$s.

\chapter{Conclusion}

Building on \cite{explanations}, \cite{rigorous} argue that explanations are necessary
when our models have a certain \textit{incompleteness}.
In the context of this thesis, we have considered the kind of incompleteness
that results from ambiguity in a dataset. When datasets do not fully specify
their decision boundaries, we have freedom to learn many models that
could accurately classify them.\footnote{Finding a way to quantify the degree of freedom a dataset affords its classifiers is an interesting topic for future research.} We generate explanations to determine which model we learned; we regularize explanations to \textit{choose} which model to learn.

In Chapter \ref{ch:1}, we provided our first formulation of explanation regularization using input gradients, a
particular kind of local explanation that is low-level but faithful and
differentiable. We then used this method to solve classification
problems despite the presence of challenging confounding factors, find diverse
solutions, and achieve high accuracy with significantly fewer training
examples. In Chapter \ref{ch:2}, we applied explanation regularization
to CNN saliency maps and showed we could make our CNNs both more interpretable and more robust
to adversarial attacks. In both chapters, we found that explanations were reliable
indicators of generalizability. In Chapter \ref{ch:3}, we reviewed other
potential ways of implementing explanation regularization, including penalties
based on alternative gradient penalties, complex surrogates, nearest neighbors,
and even abstract concepts (realized via disentangled representations).
Finally, we presented a vision for an interface that would allow users and
machine learning models to explain predictions to \textit{each other}, using a common
conceptual framework they define together.

One methodological point we would like to stress before closing is about the
importance of ground truth in evaluating explanation techniques.
Throughout this thesis, we tried to ensure that whenever we evaluated
explanations, we also had some way of
determining what those explanations \textit{should} be. For example, in Chapter \ref{ch:1},
we knew which features in our synthetic dataset actually mattered for prediction,
and we knew that models fooled by decoy datasets were sensitive to the corresponding confounds.
Regardless of whether
explanations are used in human-, application-, or functionally-grounded tasks \citep{rigorous}, having ground truth is
critical -- especially given recent criticisms of many explanation methods for
being unhelpful in detecting generalization issues
\citep{chandrasekaran2017takes}, being sensitive to meaningless changes to models
\citep{kindermans2017reliability,ghorbani2017interpretation}, or being
invariant to meaningful changes to models \citep{adebayo2018local}.
Otherwise, we may end up rationalizing our predictions rather than explaining them.

Machine learning is being deployed in more and more critical domains,
including the automotive industry, medicine, lending, bail determination
\citep{propub}, and even child maltreatment hotlines
\citep{chouldechova18a}. As we move forward, it is essential that we develop
better diagnostic tools for understanding when our models are wrong and how to right them. The promise of machine learning to improve human
lives is great, but so is its peril. To use machine learning
models responsibly, regardless of whether they seem right, we must have methods of making sure they are reasonable.

\begin{singlespacing}
  \renewcommand{\bibname}{References}
  \bibliographystyle{ecca}
  \bibliography{references}
\end{singlespacing}

\end{document}

%% file: intro.tex
The motivation for this thesis is easiest to express with a story:

\begin{displayquote}
A father decides to teach his young son what a sports car is.  Finding it
difficult to explain in words, he decides to give some examples. They stand on
a motorway bridge and as each car passes underneath, the father cries out
``that’s a sports car!'' when a sports car passes by. After ten minutes, the
father asks his son if he's understood what a sports car is. The son says,
``sure, it’s easy''. An old red VW Beetle passes by, and the son shouts –
``that’s a sports car!''. Dejected, the father asks – ``why do you say that?''.
``Because all sports cars are red!'', replies the son. \citep{barber2012bayesian}
\end{displayquote}

\noindent There is another popular version that pokes fun at the Department of
Defense:

\begin{displayquote}
In the early days of the perceptron the army decided to train an artificial
neural network to recognize tanks partly hidden behind trees in the woods. They
took a number of pictures of a woods without tanks, and then pictures of the
same woods with tanks clearly sticking out from behind trees. They then trained
a net to discriminate the two classes of pictures. The results were impressive,
and the army was even more impressed when it turned out that the net could
generalize its knowledge to pictures from each set that had not been used in
training the net. Just to make sure that the net had indeed learned to
recognize partially hidden tanks, however, the researchers took some more
pictures in the same woods and showed them to the trained net. They were
shocked and depressed to find that with the new pictures the net totally failed
to discriminate between pictures of trees with partially concealed tanks behind
them and just plain trees. The mystery was finally solved when someone noticed
that the training pictures of the woods without tanks were taken on a cloudy
day, whereas those with tanks were taken on a sunny day. The net had learned to
recognize and generalize the difference between a woods with and without
shadows! \citep{dreyfus1992artificial}
\end{displayquote}

\noindent The first story is a parable and the second is apocryphal\footnote{https://www.gwern.net/Tanks},
but both illustrate an inherent limitation in
learning by example, which is how we currently train machine learning systems: we only
provide them with inputs (questions, $X$) and outputs (answers, $y$). When we
train people to perform tasks, however, we usually provide them with
\textit{explanations}, since without them many problems are ambiguous. In machine
learning, model developers usually circumvent such ambiguities via
regularization, inductive biases (e.g. using CNNs when you need
translational invariance), or simply acquiring vast quantities of data (such
that the problem eventually becomes unambiguous, as if the child had seen
every car in the world). But there is still a risk our models will be right for
the wrong reasons -- which means that if conditions change, they will simply be wrong.

As we begin to use ML in sensitive domains such as healthcare,
this risk has highlighted the need for \textit{interpretable} models, as this final story illustrates:

\begin{displayquote}
Although models based on rules were not as accurate as the neural net models,
they were intelligible, i.e., interpretable by humans. On one of the pneumonia
datasets, the rule-based system learned the rule ``HasAsthma(x) $\to$
LowerRisk(x)'', i.e., that patients with pneumonia who have a history of asthma
have lower risk of dying from pneumonia than the general population. Needless
to say, this rule is counterintuitive. But it reflected a true pattern in the
training data: patients with a history of asthma who presented with pneumonia
usually were admitted not only to the hospital but directly to the ICU
(Intensive Care Unit).  The good news is that the aggressive care received by
asthmatic pneumonia patients was so effective that it lowered their risk of
dying from pneumonia compared to the general population. The bad news is that
because the prognosis for these patients is better than average, models trained
on the data incorrectly learn that asthma lowers risk, when in fact asthmatics
have much higher risk (if not hospitalized). \citep{caruana2015intelligible}
\end{displayquote}

\noindent In this case, a pneumonia risk prediction model learned an unhelpful rule because its training
outcomes didn't actually represent medical risk. Had the model been put into
production, it would have endangered lives. The fact that they used an
interpretable model let them realize and avoid this danger (by not using
the neural network at all). But clearly, the dataset they used still contains
information that, say, a human analyst could use to draw useful conclusions about how to treat pneumonia.
How can machine learning models utilize it despite its flaws?

This thesis seeks to provide both concrete methods for addressing
these types of problems in specific cases and more abstract arguments
about how they should be solved in general. The main strategy we will
consider is \textit{explanation regularization}, which means jointly
optimizing a machine learning model to make correct predictions
and to explain those predictions well. Quantifying the quality of an
explanation may seem difficult (especially if we would like it
to be differentiable), but we will delve into cases where it is
straightforward and intuitive, as well as strategies for making it so.

\section{Contributions}

The major contributions of this thesis are as follows:

\begin{itemize}
\item
  It presents a framework for encoding domain knowledge about a
  classification problem as local penalties on the gradient of the model's decision surface,
  which can be incorporated into the loss function of any
  differentiable model (e.g. a neural network). Applying this framework in both
  supervised and unsupervised formulations, it 
  trains models that generalize to test data from different distributions,
  which would otherwise be unobtainable by traditional optimization methods.
  (Chapter \ref{ch:1})
\item
  It applies a special case of this framework (where explanations are regularized to be simple)
  to the problem of defending against
  adversarial examples. It demonstrates increased robustness of regularized
  models to white- and black-box attacks, at a level comparable or better than
  adversarial training. It also demonstrates both increased transferability and
  \textit{interpretability} of adversarial examples created to fool regularized models,
  which we evaluate in a human subject experiment.
  (Chapter \ref{ch:2})
\item It considers cases where we can meaningfully change what
  models learn by regularizing more general types of explanations. We review literature
  and suggest directions for explanation regularization,
  using sparse gradients, input Hessians, decision trees, nearest neighbors,
  and even abstract concepts that emerge or that we encourage to emerge in deep
  neural networks. It concludes by outlining an interface for interpretable
  machine teaching.
  (Chapter \ref{ch:3})
\end{itemize}

%% file: chapter1.tex
\section{Introduction}

High-dimensional real-world datasets are often full of
ambiguities. When we train classifiers on such data, it is frequently
possible to achieve high accuracy using classifiers with qualitatively
different decision boundaries. To narrow down our choices and
encourage robustness, we usually employ regularization techniques
(e.g. encouraging sparsity or small parameter values).  We also
structure our models to ensure domain-specific invariances (e.g. using
convolutional neural nets when we would like the model to be invariant
to spatial transformations).  However, these solutions do not address
situations in which our training dataset contains subtle confounds or
differs qualitatively from our test dataset.  In these cases, our
model may fail to generalize no matter how well it is tuned.

Such generalization gaps are of particular concern for uninterpretable models
such as neural networks, especially in sensitive domains. For example,
\cite{caruana2015intelligible} describe a model intended to prioritize care
for patients with pneumonia.  The model was trained to predict hospital
readmission risk using a dataset containing attributes of patients hospitalized at least once for pneumonia.  Counterintuitively, the model learned that the
presence of asthma was a \textit{negative} predictor of readmission, when in
reality pneumonia patients with asthma are at a greater medical risk.  This model would have presented a grave safety risk if used in production.  This
problem occurred because the outcomes in the dataset reflected not just the
severity of patients' diseases but the quality of care they initially received,
which was higher for patients with asthma.

This case and others like it have motivated recent work in interpretable
machine learning, where algorithms provide explanations for domain experts to
inspect for correctness before trusting model predictions.  However, there has
been limited work in optimizing models to find not just the right prediction
but also the \textit{right explanation}.  Toward this end, this work makes the
following contributions:

\begin{itemize}
  \item We confirm empirically on several datasets that input gradient
    explanations match state of the art sample-based explanations (e.g. LIME,
    \cite{limegithub}).
  \item Given annotations about incorrect explanations for particular
    inputs, we efficiently optimize the classifier to learn alternate
    explanations (to be right for better reasons).
  \item When annotations are not available, we sequentially discover
    classifiers with similar accuracies but qualitatively different
    decision boundaries for domain experts to inspect for validity.
\end{itemize}

\subsection{Related Work}

We first define several important terms in interpretable machine learning.  All
classifiers have \textit{implicit decision rules} for converting an input into
a decision, though these rules may be opaque.  A model is
\textit{interpretable} if it provides explanations for its predictions in a
form humans can understand; an \textit{explanation} provides reliable
information about the model's implicit decision rules for a given prediction.
In contrast, we say a machine learning model is \textit{accurate} if most of
its predictions are correct, but only \textit{right for the right reasons} if
the implicit rules it has learned generalize well and conform to domain
experts' knowledge about the problem.

Explanations can take many forms \citep{explanations} and evaluating
the quality of explanations or the interpretability of a model is
difficult \citep{interpretability-mythos,rigorous}. However, within the machine
learning community recently there has been convergence
\citep{unexpected-unity} around local counterfactual explanations,
where we show how perturbing an input $x$ in various ways will affect
the model's prediction $\hat{y}$. This approach to explanations can be
domain- and model-specific (e.g. ``annotator rationales'' used to
explain text classifications by
\cite{erasure,rationalizing,rationalenetworks}).
Alternatively, explanations can be model-agnostic and
relatively domain-general, as exemplified by LIME (Local Interpretable
Model-agnostic Explanations, \cite{lime,programs-as-explanations}) which
trains and presents local sparse models of how predictions change
when inputs are perturbed.

\begin{figure}[h]
\begin{center}
\includegraphics[width=0.67\textwidth]{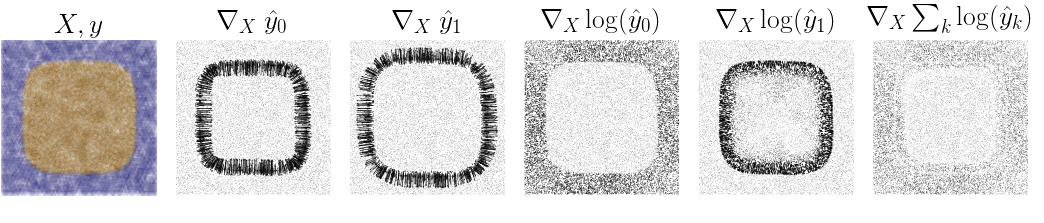} \\
\includegraphics[width=0.67\textwidth]{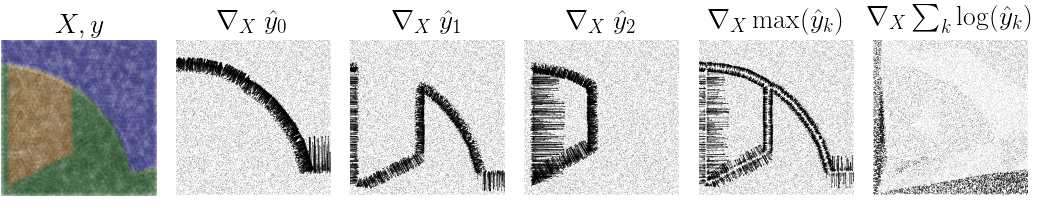}
\end{center}
\caption{Input gradients lie normal to the model's decision boundary. Examples
  above are for simple, 2D, two- and three-class datasets, with input gradients
  taken with respect to a two hidden layer multilayer perceptron with ReLU
  activations. Probability input gradients are sharpest near decision
boundaries, while log probability input gradients are more consistent within
decision regions. The sum of log probability gradients contains information
about the full model.}
\label{fig:simple-2d-explanations}
\end{figure}

The per-example perturbing and fitting process used in models such as LIME can be
computationally prohibitive, especially if we seek to explain an entire dataset
during each training iteration.  If the underlying model is differentiable,
one alternative is to use input gradients as local explanations
(\cite{baehrens2010explain} provides a particularly good introduction; see
also
\cite{grad-cam,convnets-input-gradients,text-input-gradients,input-gradients2}).
The idea is simple: the gradients of the model's output probabilities with
respect to its inputs literally describe the model's decision boundary (see
Figure~\ref{fig:simple-2d-explanations}). They are similar in spirit to the
local linear explanations of LIME but much faster to compute.

Input gradient explanations are not perfect for all use-cases---for points far
from the decision boundary, they can be uniformatively small and
do not always capture the idea of salience (see discussion and alternatives
proposed by
\cite{deeplift,layerwise,deeptaylor,integrated-gradients,fong2017interpretable}).
However, they are exactly what is required for constraining the decision
boundary.  In the past, \cite{doublebackprop} showed that applying penalties
to input gradient magnitudes can improve generalization; to our knowledge, our
application of input gradients to constrain explanations and find alternate
explanations is novel.

More broadly, none of the works above on interpretable machine
learning attempt to optimize explanations for correctness.  For SVMs
and specific text classification architectures, there exists work on
incorporating human input into decision boundaries in the form of
annotator rationales \citep{zaidan,donahue,rationalenetworks}.
Unlike our approach, these works are either tailored to specific domains
or do not fully close the loop between generating explanations and
constraining them.

\subsection{Background: Input Gradient Explanations}

Consider a differentiable model $f$ parametrized by $\theta$ with
inputs $X \inR{N}{D}$ and probability vector outputs $f(X|\theta) =
\hat{y} \inR{N}{K}$ corresponding to one-hot labels $y
\inR{N}{K}$. Its \textit{input gradient} is given by $f_X(X_n|\theta)$ or
$\nabla_X \hat{y}_n,$ which is a vector normal to the model's decision boundary
at $X_n$ and thus serves as a first-order description of the model's
behavior near $X_n$. The gradient has the same shape as each vector
$X_n$; large-magnitude values of the input gradient
indicate elements of $X_n$ that would affect $\hat{y}$ if changed.  We
can visualize explanations by highlighting portions of $X_n$ in
locations with high input gradient magnitudes.


\section{Our Approach}

We wish to develop a method to train models that are right for the right
reasons. If explanations faithfully describe a model's underlying behavior,
then constraining its explanations to match domain knowledge should cause its
underlying behavior to more closely match that knowledge too. We first describe
how input gradient-based explanations lend themselves to efficient optimization
for correct explanations in the presence of domain knowledge, and then describe
how they can be used to efficiently search for qualitatively different decision
boundaries when such knowledge is not available.

\subsection{Loss Functions that Constrain Explanations} \label{sec:loss}

When constraining input gradient explanations, there are two basic options: we
can either constrain them to be large in relevant areas or small in irrelevant
areas. However, because input gradients for relevant inputs in many models
\textit{should} be small far from the decision boundary, and because we do not
know in advance how large they should be, we opt to shrink irrelevant gradients
instead.

Formally, we define an annotation matrix $A \in \{0,1\}^{N \times D}$, which
are binary masks indicating whether dimension $d$ should be irrelevant for
predicting observation $n$. We would like $\nabla_X \hat{y}$ to be near
$0$ at these locations.  To that end, we optimize a loss function $L(\theta, X,
y, A)$ of the form \[
  L(\theta, X, y, A) = \underbrace{\sum_{n=1}^N\sum_{k=1}^K -y_{nk} \log(\hat{y}_{nk})}_{\text{Right answers}} + \underbrace{\lambda_1\sum_{n=1}^N\sum_{d=1}^D \left(A_{nd} \frac{\partial}{\partial x_{nd}} \sum_{k=1}^K  \log(\hat{y}_{nk}) \right)^2}_{\text{Right reasons}} + \underbrace{\lambda_2 \sum_{i} \theta_i^2}_{\text{Regular}},
\]
which contains familiar cross entropy and $\theta$ regularization terms along
with a new regularization term that discourages the input gradient from being
large in regions marked by $A$. This term has a regularization parameter
$\lambda_1$ which should be set such that the ``right answers'' and ``right
reasons'' terms have similar orders of magnitude; see Appendix
\ref{sec:crossval} for more details. 
Note that this loss penalizes the gradient of the
\textit{log} probability, which performed best in
practice, though in many visualizations we show $f_X$, which is the gradient of the predicted probability itself. Summing across classes led to slightly more stable results than
using the predicted class log probability $\max \log(\hat{y}_k)$, perhaps due
to discontinuities near the decision boundary (though both methods were
comparable). We did not explore regularizing input gradients of specific class
probabilities, though this would be a natural extension.

Because this loss function is differentiable with respect to $\theta$, we can
easily optimize it with gradient-based optimization methods.  We do not need
annotations (nonzero $A_n$) for every input in $X$, and in the case $A =
0^{N \times D}$, the explanation term has no effect on the loss. At the other
extreme, when $A$ is a matrix of all 1s, it encourages the model to have small
gradients with respect to its inputs; this can improve generalization on its own
\citep{doublebackprop}. Between those extremes, it biases our model against
\textit{particular} implicit rules.

This penalization approach enjoys several desirable properties.
Alternatives that specify a single $A_{d}$ for all examples presuppose
a coherent notion of global feature importance, but when
decision boundaries are nonlinear many features are only relevant in the
context of specific examples.  Alternatives that simulate
perturbations to entries known to be irrelevant (or to determine
relevance as in \cite{lime}) require defining domain-specific
perturbation logic; our approach does not.  Alternatives that apply
hard constraints or completely remove elements identified by
$A_{nd}$ miss the fact that the entries in $A$ may be
imprecise even if they are human-provided.  Thus, we opt to preserve
potentially misleading features but softly penalize their use.

\subsection{Find-Another-Explanation: Discovering Many Possible Rules without Annotations} \label{sec:fae}

Although we can obtain the annotations $A$ via experts as in \cite{zaidan}, we
may not always have this extra information or know the ``right reasons.'' In
these cases, we propose an approach that iteratively adapts $A$ to discover
multiple models accurate for \emph{qualitatively different} reasons; a domain
expert could then examine them to determine which is the right for the best
reasons.  Specifically, we generate a ``spectrum'' of models with different
decision boundaries by iteratively training models, explaining $X$, then
training the next model to differ
from previous iterations: \[
\begin{aligned}
  A_0 & = 0,                                      & \theta_0 = \argmin_{\theta} L(\theta, X, y, A_0), \\
  A_1 & = M_c\left[f_X|\theta_0\right],           & \theta_1 = \argmin_{\theta} L(\theta, X, y, A_1), \\
A_2 & = M_c\left[f_X|\theta_1\right] \cup A_1,\ & \theta_2 = \argmin_{\theta} L(\theta, X, y, A_2), \\
\end{aligned}
\]
\noindent \begin{center}$\hdots$\end{center}
where the function $M_c$ returns a binary mask indicating which gradient
components have a magnitude ratio (their magnitude divided by the largest
component magnitude) of at least $c$ and where we abbreviated the
input gradients of the entire training set $X$ at $\theta_i$ as
$f_X|\theta_i$. In other words, we regularize input gradients where they were
largest in magnitude
previously.  If, after repeated iterations, accuracy decreases or
explanations stop changing (or only change after significantly increasing
$\lambda_1$), then we may have spanned the space of possible models.\footnote{Though one can design simple pathological cases where we do not discover all models with this method; we explore an alternative version in Appendix \ref{sec:fae2} that addresses some of these cases.}  All of the
resulting models will be accurate, but for different reasons; although we do
not know which reasons are best, we can present them to a domain expert for
inspection and selection. We can also prioritize labeling or reviewing examples
about which the ensemble disagrees. Finally, the size of the ensemble
provides a rough measure of dataset redundancy. 

\section{Empirical Evaluation}

We demonstrate explanation generation, explanation constraints, and the
find-another-explanation method on a toy color dataset and three real-world
datasets. In all cases, we used a multilayer perceptron with two hidden layers
of size 50 and 30, ReLU nonlinearities with a softmax output, and a $\lambda_2
= 0.0001$ penalty on $\norm{\theta}_2^2$.  We trained the network using Adam
\citep{adam} with a batch size of 256 and Autograd \citep{autograd}.  For
most experiments, we used an explanation L2 penalty of $\lambda_1 = 1000$,
which gave our ``right answers'' and ``right reasons'' loss terms similar
magnitudes. More details about cross-validation are included in
Appendix~\ref{sec:crossval}.  For the cutoff value $c$ described in Section
\ref{sec:fae} and used for display, we often chose 0.67, which tended to
preserve 2-5\% of gradient components (the average number of qualifying
elements tended to fall exponentially with $c$).  Code for all experiments is
available at
\texttt{\href{https://github.com/dtak/rrr}{https://github.com/dtak/rrr}}.


\subsection{Toy Color Dataset}

We created a toy dataset of $5 \times 5 \times 3$ RGB images with four possible
colors.  Images fell into two classes with two independent
decision rules a model could implicitly learn: whether their four corner pixels
were all the same color, and whether their top-middle three pixels were all different
colors. Images in class 1 satisfied both conditions and images in class 2
satisfied neither.  Because only corner and top-row pixels are relevant, we
expect any faithful explanation of an accurate model to highlight them.

\begin{figure}[h]
  \centering
  \includegraphics[width=0.5\textwidth]{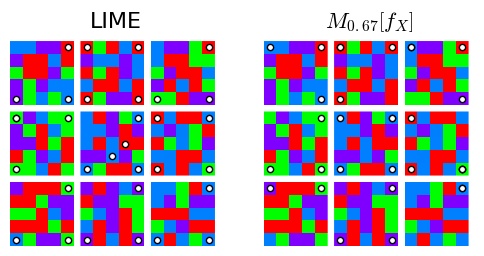}
  \caption{Gradient vs. LIME explanations of nine perceptron predictions on the
    Toy Color dataset. For gradients, we plot dots above
    pixels identified by $M_{0.67}\left[f_X\right]$ (the top 33\%
  largest-magnitude input gradients), and for LIME, we select the top 6
features (up to 3 can reside in the same RGB pixel). Both methods suggest that
the model learns the corner rule.}
  \label{fig:colors-vs-lime}
\end{figure}

In Figure \ref{fig:colors-vs-lime}, we see both LIME and input
gradients identify the same relevant pixels, which suggests that (1)
both methods are effective at explaining model predictions, and (2)
the model has learned the corner rather than the top-middle rule,
which it did consistently across random restarts.

\begin{figure}[h]
\centering
\includegraphics[width=0.67\textwidth]{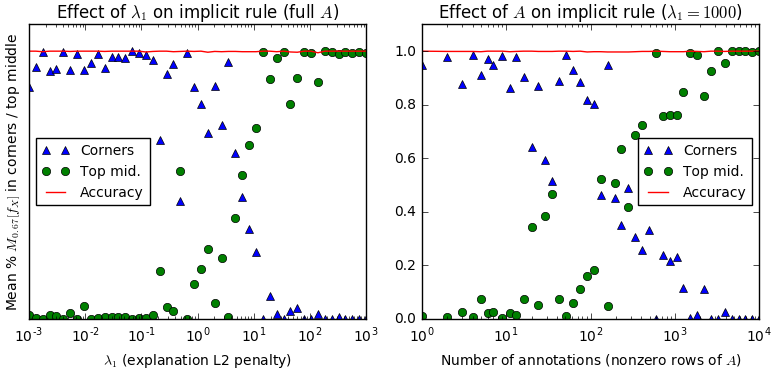}
\caption{Implicit rule transitions as we increase $\lambda_1$ and the number of
  nonzero rows of $A$. Pairs of points represent the fraction of
  large-magnitude ($c=0.67$) gradient components in the corners and top-middle
  for 1000 test examples, which almost always add to 1 (indicating the
  model is most sensitive to these elements alone, even during
transitions). Note there is a wide regime where the model learns a
hybrid of both rules.}
\label{fig:color-transitions}
\end{figure}

\begin{figure}[h]
\centering
\includegraphics[width=0.67\textwidth]{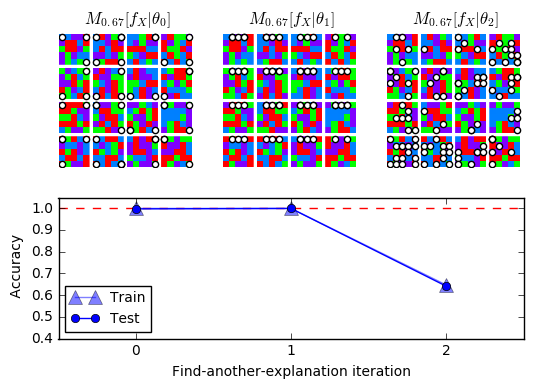}
\caption{Rule discovery using find-another-explanation method with 0.67 cutoff
and $\lambda_1=10^3$ for $\theta_1$ and $\lambda_1=10^6$ for $\theta_2$. Note
how the first two iterations produce explanations corresponding to the two
rules in the dataset while the third produces very noisy explanations (with low
accuracies).}
\label{fig:color-fae}
\end{figure}

However, if we train our model with a nonzero $A$ (specifically, setting
$A_{nd}=1$ for corners $d$ across examples $n$), we were able to cause it to
use the other rule.  Figure \ref{fig:color-transitions} shows how the model
transitions between rules as we vary $\lambda_1$ and the number of examples
penalized by $A$.  This result demonstrates that the model can be made to learn
multiple rules despite only one being commonly reached via standard
gradient-based optimization methods. However, it depends on knowing a good
setting for $A$, which in this case would still require annotating on the order
of $10^3$ examples, or 5\% of our dataset (although always including examples
with annotations in Adam minibatches let us consistently switch rules with
only 50 examples, or 0.2\% of the dataset).

Finally, Figure \ref{fig:color-fae} shows we can use the
find-another-explanation technique from Sec.~\ref{sec:fae} to discover the
other rule without being given $A$.  Because only two rules lead to high
accuracy on the test set, the model performs no better than random guessing
when prevented from using either one (although we have to increase the penalty
high enough that this accuracy number may be misleading - the essential point
is that after the first iteration, explanations stop changing).  Lastly, though
not directly relevant to the discussion on interpretability and explanation, we
demonstrate the potential of explanations to reduce the amount of data required
for training in Appendix~\ref{sec:learnfast}.


\subsection{Real-world Datasets}

To demonstrate real-world, cross-domain applicability, we test our approach on
variants of three familiar machine learning text, image, and tabular datasets:

\begin{itemize}
\item \textbf{20 Newsgroups:} As in \cite{lime}, we test input
  gradients on the \texttt{alt.atheism}
  vs. \texttt{soc.religion.christian} subset of the 20 Newsgroups
  dataset \cite{uciml}. We used the same two-hidden layer network
  architecture with a TF-IDF vectorizer with 5000 components, which
  gave us a 94\% accurate model for $A=0$.

\item \textbf{Iris-Cancer:} We concatenated all examples in classes 1 and 2
  from the Iris dataset with the the first 50 examples from each class in the
  Breast Cancer Wisconsin dataset \citep{uciml} to create a composite dataset
  $X\inR{100}{34},y\in\{0,1\}$. Despite the dataset's small size, our network
  still obtains an average test accuracy of 92\% across 350 random
  $\frac{2}{3}$-$\frac{1}{3}$ training-test splits.  However, when we modify
  our test set to remove the 4 Iris components, average test accuracy falls to
  81\% with higher variance, suggesting the model learns to depend on Iris
  features and suffers without them.  We verify that our explanations reveal
  this dependency and that regularizing them avoids it.

\item \textbf{Decoy MNIST:} On the baseline MNST dataset \citep{mnist}, our
  network obtains 98\% train and 96\% test accuracy. However, in Decoy MNIST,
  images $x$ have $4 \times 4$ gray swatches in randomly chosen corners whose
  shades are functions of their digits $y$ in training (in particular, $255 -
  25y$) but are random in test. On this dataset, our model has a higher 99.6\%
  train accuracy but a much lower 55\% test accuracy, indicating that the decoy
  rule misleads it. We verify that both gradient and LIME explanations let
  users detect this issue and that explanation regularization lets us
  overcome it.
\end{itemize}

\begin{figure}[h]
	\begin{center}
	\includegraphics[width=\textwidth]{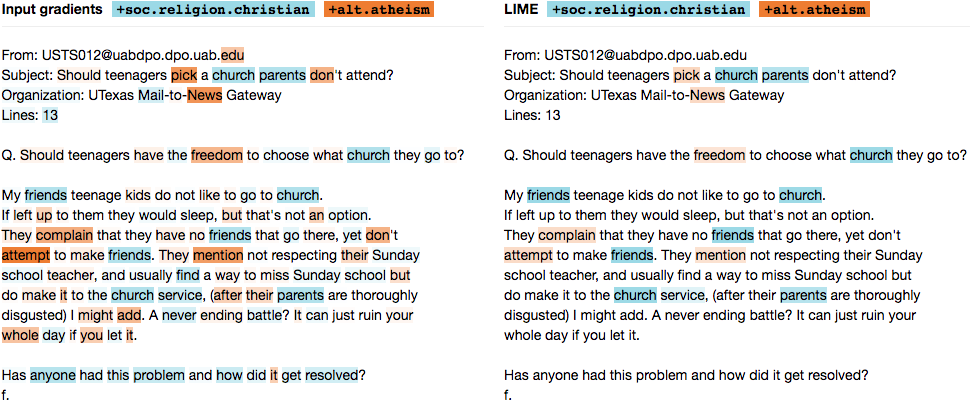}
	\end{center}
  \caption{Words identified by LIME vs. gradients on an example from the atheism vs.
    Christianity subset of 20 Newsgroups. More
    examples are available at
    {\href{https://github.com/dtak/rrr}{https://github.com/dtak/rrr}}.
    Words are blue if they support {soc.religion.christian}
    and orange if they support {alt.atheism}, with opacity
    equal to the ratio of the magnitude of the word's
    weight to the largest magnitude weight. LIME generates sparser explanations
    but the weights and signs of terms identified by both methods match
    closely. Note that both methods reveal some aspects of the model that
    are intuitive (``church'' and ``service'' are associated with Christianity), some aspects that are not (``13'' is associated with
    Christianity, ``edu'' with atheism), and some that are debatable (``freedom''
    is associated with atheism, ``friends'' with Christianity).}
	\label{fig:20ng-vs-lime}
\end{figure}

\begin{figure}[h]
	\begin{center}
	\includegraphics[width=0.67\textwidth]{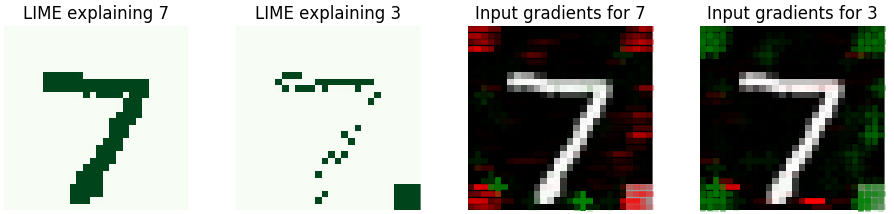}
	\end{center}
  \caption{Input gradient explanations for Decoy MNIST vs. LIME, using the LIME
  image library \protect\cite{limegithub}. In this example, the
  model incorrectly predicts 3 rather than 7 because of the decoy swatch.}
	\label{fig:mnist-vs-lime}
\end{figure}

\begin{figure}[h]
	\begin{center}
	\includegraphics[width=0.67\textwidth]{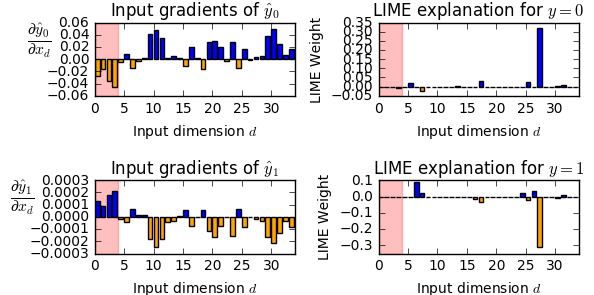}
	\end{center}
  \caption{Iris-Cancer features identified by input gradients vs. LIME, with
  Iris features highlighted in red. Input gradient explanations are more
faithful to the model. Note that most gradients change sign
when switching between $\hat{y}_0$ and $\hat{y}_1$, and that
the magnitudes of input gradients are different across examples,
which provides information about examples' proximity to the decision boundary.}
	\label{fig:iris-vs-lime}
\end{figure}

\noindent \textbf{Input gradients are consistent with sample-based methods such as
LIME, and faster.} On 20 Newsgroups (Figure \ref{fig:20ng-vs-lime}), input
gradients are less sparse but identify all of the same words in the document
with similar weights. Note that input gradients also identify words outside the
document that would affect the prediction if added.

\begin{figure}[h]
	\begin{center}
	\includegraphics[width=0.49\textwidth]{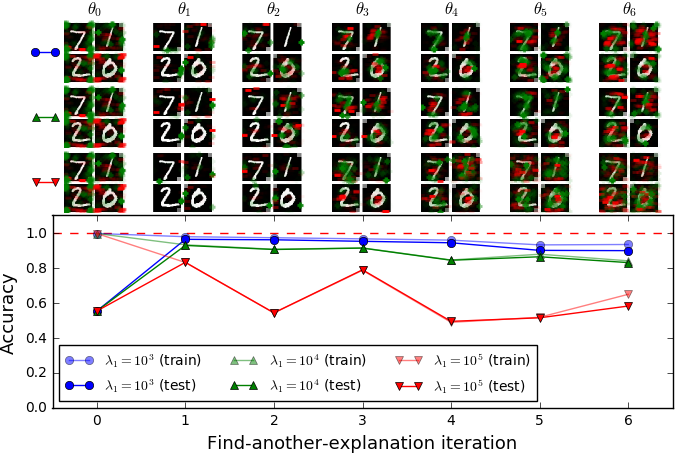}
	\includegraphics[width=0.49\textwidth]{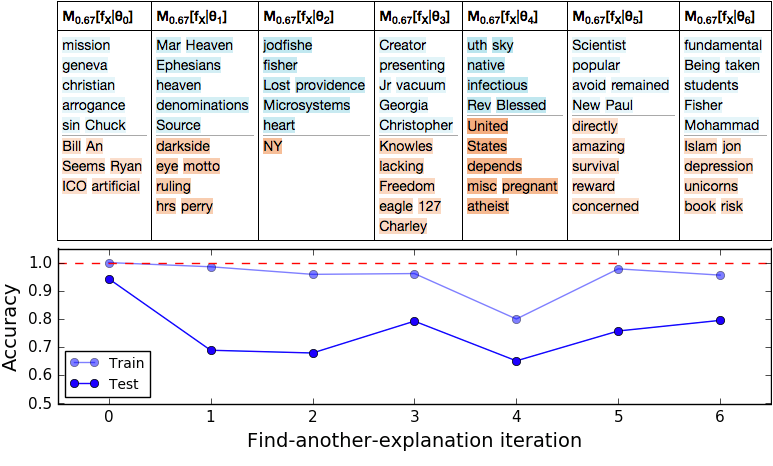}
	\includegraphics[width=0.49\textwidth]{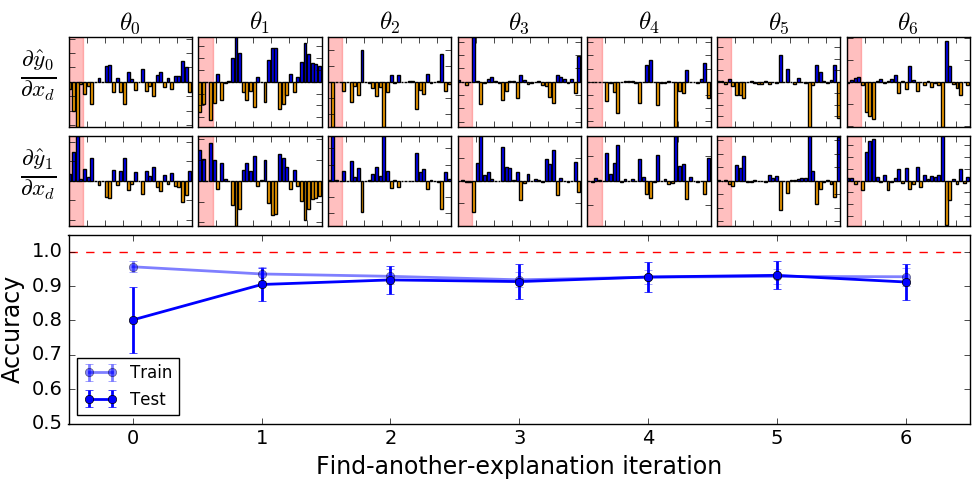}
	\end{center}
  \caption{Find-another-explanation results on Iris-Cancer (top; errorbars show
    standard deviations across 50 trials), 20 Newsgroups (middle; blue supports
    Christianity and orange supports atheism, word opacity set to magnitude
    ratio), and Decoy MNIST (bottom, for three values of $\lambda_1$ with
    scatter opacity set to magnitude ratio cubed). Real-world datasets are
    often highly redundant and allow for diverse models with similar
  accuracies. On Iris-Cancer and Decoy MNIST, both explanations and accuracy
results indicate we overcome confounds after 1-2 iterations without any prior
knowledge about them encoded in $A$.}
	\label{fig:fae}
\end{figure}

On Decoy MNIST (Figure \ref{fig:mnist-vs-lime}), both LIME and input gradients
reveal that the model predicts 3 rather than 7 due to the color
swatch in the corner. Because of their fine-grained resolution, input gradients
sometimes better capture counterfactual behavior, where extending or adding
lines outside of the digit to either reinforce it or transform it into another
digit would change the predicted probability (see also Figure \ref{fig:fae}).
LIME, on the other hand, better captures the fact that the main portion of the
digit is salient (because its super-pixel perturbations add and remove larger
chunks of the digit).

On Iris-Cancer (Figure \ref{fig:iris-vs-lime}), input gradients actually
outperform LIME.  We know from the accuracy difference that Iris features are
important to the model's prediction, but LIME only identifies a single
important feature, which is from the Breast Cancer dataset (even when we vary
its perturbation strategy).  This example, which is tabular and contains
continuously valued rather categorical features, may represent a pathological
case for LIME, which operates best when it can selectively mask a small number
of meaningful chunks of its inputs to generate perturbed samples.  For truly
continuous inputs, it should not be surprising that explanations based on
gradients perform best.

There are a few other advantages input gradients have over
sample-based perturbation methods. On 20 Newsgroups, we noticed that
for very long documents, explanations generated by the sample-based
method LIME are often overly sparse,
and there are many words identified as significant by input gradients
that LIME ignores. This may be because the number of features LIME
selects must be passed in as a parameter beforehand, and it may also
be because LIME only samples a fixed number of times. For sufficiently
long documents, it is unlikely that sample-based approaches will mask
every word even once, meaning that the output becomes increasingly
nondeterministic---an undesirable quality for explanations. To resolve
this issue, one could increase the number of samples, but that would
increase the computational cost since the model must be evalutated at
least once per sample to fit a local surrogate.  Input gradients, on
the other hand, only require on the order of one model evaluation
\textit{total} to generate an explanation of similar quality
(generating gradients is similar in complexity to predicting
probabilities), and furthermore, this complexity is based on the
vector length, \textit{not} the document length. This issue
(underscored by Table \ref{tab:lime-vs-grad-performance}) highlights
some inherent scalability advantages input gradients enjoy over
sample-based perturbation methods.

\begin{table}[h]
\centering
\begin{tabular}{|c|c|c|c|} \hline
    & LIME & Gradients & Dimension of $x$ \\
\hline
Iris-Cancer & 0.03s & 0.000019s & 34 \\
\hline
Toy Colors    & 1.03s  & 0.000013s & 75 \\
\hline
Decoy MNIST & 1.54s  & 0.000045s & 784 \\
\hline
20 Newsgroups & 2.59s  & 0.000520s & 5000 \\
\hline
\end{tabular}
\caption{Gradient vs. LIME runtimes per explanation. Note that each method uses
  a different version of LIME; Iris-Cancer and Toy Colors use
  \texttt{lime\_tabular} with continuous and quartile-discrete perturbation
  methods, respectively, Decoy MNIST uses \texttt{lime\_image}, and 20
Newsgroups uses \texttt{lime\_text}. Code was executed on a laptop and input
gradient calculations were not optimized for performance, so runtimes are only
meant to provide a sense of scale.}

\label{tab:lime-vs-grad-performance}
\end{table}

\begin{figure}[h]
	\begin{center}
	\includegraphics[width=0.67\textwidth]{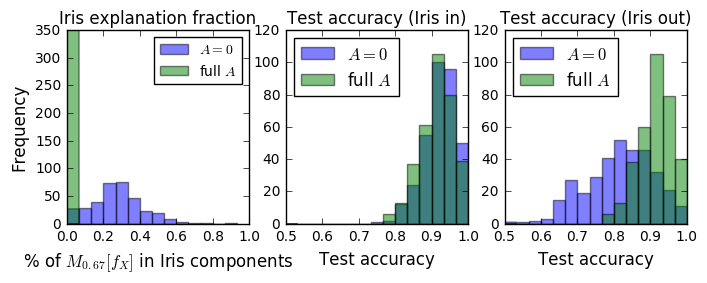}
  \end{center} \caption{Overcoming confounds using explanation constraints on
    Iris-Cancer (over 350 random train-test splits).  By default ($A=0$), input
    gradients tend to be large in Iris dimensions, which results in lower
    accuracy when Iris is removed from the test set. Models trained with
  $A_{nd}=1$ in Iris dimensions (full $A$) have almost exactly the same test
accuracy with and without Iris.}
	\label{fig:iris-confounds}
\end{figure}

\begin{figure}[h]
	\begin{center}
	\includegraphics[width=0.5\textwidth]{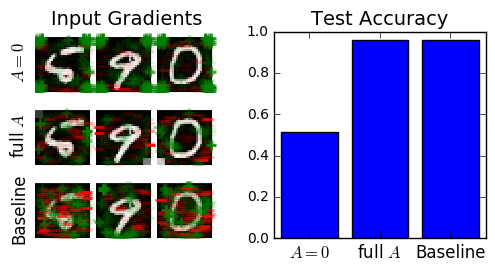}
	\end{center}
  \caption{Training with explanation constraints on Decoy MNIST. Accuracy is
    low ($A=0$) on the swatch color-randomized test set unless
    the model is trained with $A_{nd}=1$ in swatches (full $A$).
    In that case, test accuracy matches the same architecture's performance on
    the standard MNIST dataset (baseline).}
	\label{fig:mnist-confounds}
\end{figure}

$\,$ \\ \noindent \textbf{Given annotations, input gradient regularization finds solutions
consistent with domain knowledge.} Another key advantage of using an
explanation method more closely related to our model is that we can then
incorporate explanations into our training process, which are most useful when
the model faces ambiguities in how to classify inputs. We deliberately
constructed the Decoy MNIST and Iris-Cancer datasets to have this kind of
ambiguity, where a rule that works in training will not generalize to test.
When we train our network on these confounded datasets, their test accuracy is
better than random guessing, in part because the decoy rules are not simple and
the primary rules not complex, but their performance is still significantly
worse than on a baseline test set with no decoy rules. By penalizing
explanations we know to be incorrect using the loss function defined in Section
\ref{sec:loss}, we are able to recover that baseline test accuracy, which we
demonstrate in Figures \ref{fig:iris-confounds} and \ref{fig:mnist-confounds}.

$\,$ \\ \noindent \textbf{When annotations are unavailable, our
find-another-explanation method discovers diverse classifiers.} As we saw with
the Toy Color dataset, even if almost every row of $A$ is 0, we can still
benefit from explanation regularization (meaning practitioners can gradually
incorporate these penalties into their existing models without much upfront
investment).  However, annotation is never free, and in some cases we either do
not know the right explanation or cannot easily encode it.  Additionally, we
may be interested in exploring the structure of our model and dataset in a less
supervised fashion.  On real-world datasets, which are usually overdetermined,
we can use find-another-explanation to discover $\theta$s in shallower local
minima that we would normally never explore. Given enough models right for
different reasons, hopefully at least one is right for the right reasons.

Figure \ref{fig:fae} shows find-another-explanation results for our three
real-world datasets, with example explanations at each iteration above and
model train and test accuracy below. For Iris-Cancer, we find that the initial
iteration of the model heavily relies on the Iris features and has high train
but low test accuracy, while subsequent iterations have lower train
but higher test accuracy (with smaller gradients in Iris components). In other
words, we spontaneously obtain a more generalizable model without a predefined $A$
alerting us that the first four features are misleading.

Find-another-explanation also overcomes confounds on Decoy MNIST, needing only
one iteration to recover baseline accuracy. Bumping $\lambda_1$ too high (to
the point where its term is a few orders of magnitude larger than the
cross-entropy) results in more erratic behavior. Interestingly, in a process
remniscent of distillation \citep{distillation}, the gradients themselves
become more evenly and intuitively distributed at later iterations. In many
cases they indicate that the probabilities of certain digits increase when we
brighten pixels along or extend their distinctive strokes, and that they
decrease if we fill in unrelated dark areas, which seems desirable. However, by
the last iteration, we start to revert to using decoy swatches in some cases.

On 20 Newsgroups, the words most associated with \texttt{alt.atheism} and \\
\texttt{soc.religion.christian} change between iterations but remain mostly intuitive in their associations.
Train accuracy mostly remains high while test accuracy is unstable.

For all of these examples, accuracy remains high even as decision boundaries
shift significantly. This may be because real-world data tends to contain
significant redundancies.

\subsection{Limitations}

Input gradients provide faithful information about a model's rationale for a
prediction but trade interpretability for efficiency. In
particular, when input features are not individually meaningful to users (e.g.
for individual pixels or word2vec components), input gradients may be difficult to
interpret and $A$ may be difficult to specify. Additionally, because they can
be 0 far from the decision boundary, they do not capture the idea of salience
as well as other methods
\citep{visualizing-convnets,integrated-gradients,deeptaylor,layerwise,deeplift}.
However, they are necessarily faithful to the model and easy to incorporate
into its loss function. Input gradients are first-order linear approximations
of the model; we might call them first-order explanations.

\section{Discussion}

In this chapter, we showed that:
\begin{itemize}
  \item On training sets that contain confounds which would fool \textit{any} model trained just to make correct predictions, we can use gradient-based explanation regularization to learn models that still generalize to test. These results imply that gradient regularization actually changes \textit{why} our model makes predictions.
  \item When we lack expert annotations, we can still use our method in an unsupervised manner to discover models that make predictions for different reasons. This ``find-another-explanation'' technique allowed us to overcome confounds on Decoy MNIST and Iris-Cancer, and even quantify the ambiguity present in the Toy Color dataset.
  \item Input gradients are consistent with sample-based
methods such as LIME but faster to compute and sometimes more faithful to the
model, especially for continuous inputs.
\end{itemize}
Our consistent results on several diverse datasets show that input gradients merit
further investigation as building blocks for optimizable explanations; there
exist many options for further advancements such as weighted annotations $A$,
different penalty norms,
and more general specifications of whether features should be positively or
negatively predictive of specific classes for specific inputs.

Finally, our ``right for the right reasons'' approach may be of use in solving
related problems, e.g. in integrating causal inference with deep neural networks
or maintaining robustness to adversarial examples (which we discuss in Chapter \ref{ch:2}).
Building on our find-another-explanation
results, another promising direction is to let humans in the loop
interactively guide models towards correct explanations.
Overall, we feel that
developing methods of ensuring that models are right for better reasons is
essential to overcoming the inherent obstacles to generalization posed by
ambiguities in real-world datasets.

\begin{subappendices}
\section{Cross-Validation}
\label{sec:crossval}

Most regularization parameters are selected to maximize accuracy on a
validation set. However, when your training and validation sets share the same
misleading confounds, validation accuracy may not be a good proxy for test
accuracy. Instead, we recommend increasing the explanation regularization
strength $\lambda_1$ until the cross-entropy and ``right reasons'' terms have
roughly equal magnitudes (which corresponds to the region of highest test
accuracy below). Intuitively, balancing the terms in this way should push our
optimization away from cross-entropy minima that violate the explanation
constraints specified in $A$ and towards ones that correspond to ``better
reasons.'' Increasing $\lambda_1$ too much makes the cross-entropy term
negligible. In that case, our model performs no better than random guessing.

\begin{figure}[h]
\begin{center}
\includegraphics[width=0.5\textwidth]{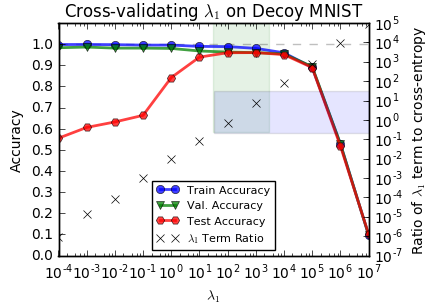}
\end{center}
\caption{\small Cross-validating $\lambda_1$. The regime of highest accuracy (highlighted) is also where the initial cross-entropy and $\lambda_1$ loss terms have similar magnitudes. Exact equality is not required; being an order of magnitude off does not significantly affect accuracy.}
\label{fig:crossval}
\end{figure}

\section{Learning with Less Data} \label{sec:learnfast}

It is natural to ask whether explanations can reduce data requirements. Here we
explore that question on the Toy Color dataset using four variants of $A$ 
(with $\lambda_1$ chosen to match loss terms at each $N$).

\begin{figure}[h]
\begin{center}
\includegraphics[width=0.67\textwidth]{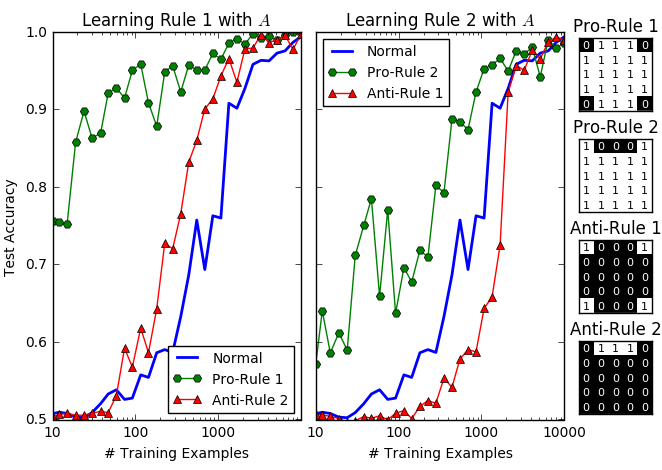}
\end{center}
  \caption{\small Explanation regularization can reduce data requirements.}
\label{fig:learnfast}
\end{figure}

\noindent We find that when $A$ is set to the Pro-Rule 1 mask, which penalizes
all pixels except the corners, we reach 95\% accuracy with fewer than 100
examples (as compared to $A=0$, where we need almost 10000). Penalizing the
top-middle pixels (Anti-Rule 2) or all pixels except the top-middle (Pro-Rule
2) also consistently improves accuracy relative to data. Penalizing the corners
(Anti-Rule 1), however, reduces accuracy until we reach a threshold $N$. This
may be because the corner pixels can match in 4 ways, while the
top-middle pixels can differ in $4\cdot3\cdot2=24$ ways, suggesting that Rule 2
could be inherently harder to learn from data and positional explanations
alone.

\section{Simultaneous Find-Another-Explanation} \label{sec:fae2}

In Section \ref{sec:fae}, we introduced a method of training
classifiers to make predictions for different reasons by
sequentially augmenting $A$ to penalize more features.
However, as our ensemble grows, $A$ can saturate to $\mathbb{1}^{N \times D}$, and subsequent models will be trained with uniform gradient regularization.
While these models may have desirable properties (which we explore in the following chapter), they will not be diverse.

\begin{figure}[b]
\includegraphics[width=0.32\textwidth]{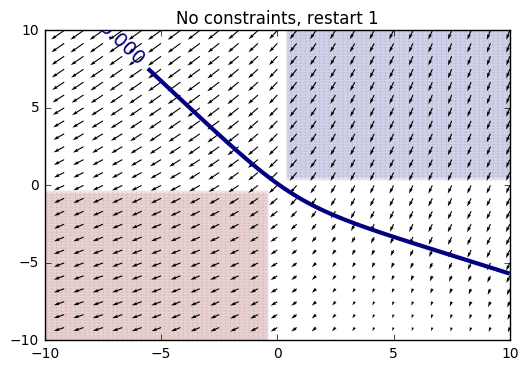}
\includegraphics[width=0.32\textwidth]{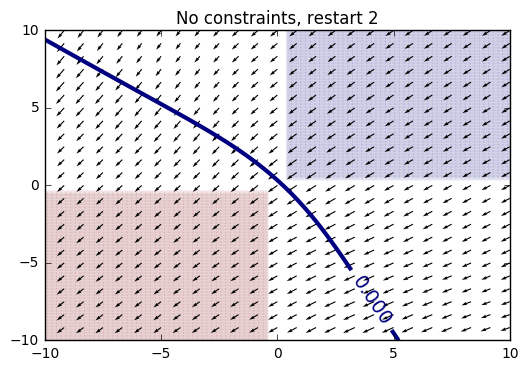}
\includegraphics[width=0.32\textwidth]{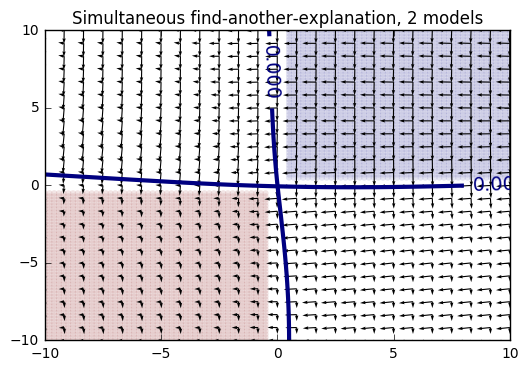}
\caption{Toy 2D problem with one degree of decision boundary freedom. Across random restarts (left two plots), we tend to learn a boundary in which both features are significant, which prevents sequential find-another-explanation from producing diverse models. If we jointly train two models with a penalty on the cosine similarity of their gradients (right plot), they end up with orthogonal boundaries.}
\label{fig:simul-fae-2d}
\end{figure}

As a simple example, consider a 2D dataset with one class
confined to the first quadrant and the other confined to the third. In theory, we have a full degree of decision freedom; it should be possible to learn two perfect and fully orthogonal boundaries (one
horizontal, one vertical). However, when we train our first MLP, it learns a diagonal surface; both features have large gradients everywhere, so $A=\mathbb{1}^{N \times 2}$ immediately. To resolve this, we propose a simultaneous training procedure: \begin{equation}
  \theta_1^*,\hdots,\theta_M^* = \argmin_{\theta_1,\hdots,\theta_M}
  \sum_{a=1}^M \mathcal{L}(y,f(X|\theta_a))
    + \sum_{a=1}^M \sum_{b=a+1}^M
    \text{\texttt{sim}}(f_X(X|\theta_a), f_X(X|\theta_b)),
\end{equation}
where $\mathcal{L}$ refers to our single-model loss function, and for our similarity measure we use the squared cosine similarity
$\text{\texttt{sim}}(v,w) = \frac{(v^Tw)^2}{(v^Tv)(w^Tw)+\epsilon}$,
where we add $\epsilon=10^{-6}$ to the denominator for numerical stability.
Squaring the cosine similarity ensures our penalty is positive, 
is minimized by orthogonal boundaries, and is soft
for nearly orthogonal boundaries. We show in Figure \ref{fig:simul-fae-2d} that this lets us obtain the two desired models.

\end{subappendices}

%% file: chapter2.tex
\section{Introduction}

In the previous chapter, we used input gradient penalties to
encourage neural networks to make predictions for specific reasons.
We demonstrated this on ``decoy'' datasets deliberately designed
to deceive models making decisions for different reasons.
This philosophy of testing -- that we should measure generalization by testing
on data from a different distribution than we trained on --
can be taken to its extreme by testing models in
an adversarial setting, where neural networks
have known vulnerabilities \citep{szegedy2013intriguing}.
In this chapter, we consider whether a domain knowledge-agnostic application of
explanation regularization (a uniform
L2 penalty on input gradients, similar in spirit to Ridge regression on the
model's local linear approximations) could help defend against adversarial
examples.

Adversarial examples pose serious obstacles for the adoption of
neural networks in settings which are
security-sensitive or have legal ramifications
\citep{kang2017prediction}.
Although many techniques
for generating these examples (which we call ``attacks'') require access to
model parameters, \citet{papernot2017practical} have shown that it is possible
and even practical to attack black-box models in the real world, in large part
because of \textit{transferability}; examples
generated to fool one model tend to fool \textit{all} models trained on the
same dataset. Particularly for images, these adversarial examples can be
constructed to fool models across a variety of scales and perspectives
\citep{athalye2017synthesizing}, which poses a problem for the adoption of deep
learning models in systems like self-driving cars.

Although there has recently been a great deal of research in adversarial
defenses, many of these methods have struggled to achieve robustness to
transferred adversarial examples \citep{tramer2017space}. Some of the most effective defenses simply detect and reject them
rather than making predictions \citep{xu2017feature}.
The most
common, ``brute force'' solution is adversarial training, where we include a mixture of
normal and adversarially-generated examples in the training set
\citep{kurakin2016adversarial}. However, \citet{tramer2017ensemble} show that
the robustness adversarial training provides can be circumvented by randomizing or transferring
perturbations from other models (though ensembling helps).

As we noted in Chapter \ref{ch:1}, domain experts are also often concerned that DNN predictions are uninterpretable.  The lack of interpretability is particularly problematic in
domains where algorithmic bias is often a factor \citep{propub} or in
medical contexts where safety risks can arise when there is mismatch between how a model is trained and used
\citep{caruana2015intelligible}. For computer vision models (the primary target of adversarial attacks), the most common class of explanation is the saliency map, either at the level of raw pixels, grid chunks, or superpixels \citep{lime}.

The local linear approximation provided by raw input gradients \citep{baehrens2010explain} is sometimes used for pixel-level saliency maps \citep{convnets-input-gradients}. However, computer vision practitioners tend not to examine raw input gradients
because they are noisy and difficult to interpret.  This issue has
spurred the development of techniques like integrated gradients
\citep{integrated-gradients} and SmoothGrad
\citep{smoothgrad} that generate smoother, more interpretable saliency maps
from noisy gradients. The rationale behind these techniques is that, while the
local behavior of the model may be noisy, examining the gradients over larger
length scales in input space provides a better intution about the model's
behavior.

However, raw input gradients are \textit{exactly} what many attacks use to
generate adversarial examples. Explanation techniques which smooth out
gradients in background pixels may be inappropriately hiding the fact that the
model is quite sensitive to them.  We consider that perhaps the need for these smoothing
techniques in the first place is indicative of a problem with our models, related to their adversarial vulnerability and capacity to overfit. Perhaps it is fundamentally hard for adversarially vulnerable models to be interpretable.

On the other hand, perhaps it is hard for interpretable models to be adversarially vulnerable.
Our hypothesis is that by training a model to have smooth input
gradients with fewer extreme values, it will not only be more interpretable but
also more resistant to adversarial examples. In the experiments that
follow we confirm this hypothesis using uniform gradient regularization,
which optimizes the model to have smooth
input gradients with respect to its predictions during training.
Using this technique, we demonstrate robustness to adversarial
examples across multiple model architectures and datasets, and in particular
demonstrate robustness to \textit{transferred} adversarial examples: gradient-regularized models maintain significantly higher accuracy on examples
generated to fool other models than baselines. Furthermore, both
qualitatively and in human subject experiments, we find that
adversarial examples generated to fool gradient-regularized models are, in a
particular sense, more ``interpretable'': they fool humans as well.

\section{Background}

In this section, we will (re)introduce notation, and give a brief overview of the
baseline attacks and defenses against which we will test and
compare our methods. The methods we will analyze again apply to all differentiable
classification models $f_\theta(X)$, which are functions parameterized by
$\theta$ that return predictions $\hat{y} \inR{N}{K}$ given inputs $X
\inR{N}{D}$. These predictions indicate the probabilities that each of $N$
inputs in $D$ dimensions belong to each of $K$ class labels. To train these
models, we try to find sets of parameters $\theta^*$ that minimize the total
information distance between the predictions $\hat{y}$ and the true labels $y$
(also $\inR{N}{K}$, one-hot encoded) on a training set: \begin{equation}
\begin{split}
  \theta^* & = \argmin_\theta \sum_{n=1}^N \sum_{k=1}^K -y_{nk} \log f_\theta(X_n)_k, \\
\end{split}
\label{eq:normal-training}
\end{equation}
which we will sometimes write as \[
  \argmin_\theta H(y, \hat{y}),
\]
with $H$ giving the sum of the cross entropies between the predictions and the
labels.

\subsection{Attacks}

\subsubsection{Fast Gradient Sign Method (FGSM)}

\citet{fgsm} introduced this first method of generating adversarial examples by
perturbing inputs in a manner that increases the local linear approximation of
the loss function:
\begin{equation}
X_{\text{FGSM}} = X + \epsilon\, \text{sign}\left(\nabla_x H(y, \hat{y})\right)
\label{eq:fgsm}
\end{equation}
If $\epsilon$ is small, these adversarial examples are indistinguishable from
normal examples to a human, but the network performs significantly worse on
them.

\citet{tgsm} noted that one can iteratively perform this attack with a small
$\epsilon$ to induce misclassifications with a smaller total perturbation (by
following the nonlinear loss function in a series of small linear steps rather
than one large linear step).

\subsubsection{Targeted Gradient Sign Method (TGSM)}

A simple modification of the Fast Gradient Sign Method is the Targeted Gradient
Sign Method, introduced by \citet{tgsm}. In this attack, we attempt to decrease
a modified version of the loss function that encourages the model to
misclassify examples in a specific way:
\begin{equation}
  X_{\text{TGSM}} = X - \epsilon\, \text{sign}\left(\nabla_x H(y_{\text{target}}, \hat{y})\right),
\label{eq:tgsm}
\end{equation}
where $y_{\text{target}}$ encodes an alternate set of labels we would like the
model to predict instead. In the digit classification experiments below, we
often picked targets by incrementing the labels $y$ by 1 (modulo 10), which we
will refer to as $y_{+1}$. The TGSM can also be performed iteratively.

\subsubsection{Jacobian-based Saliency Map Approach (JSMA)}

The final attack we consider, the Jacobian-based Saliency Map Approach (JSMA), also
takes an adversarial target vector $y_{\text{target}}$. It iteratively searches
for pixels or pairs of pixels in $X$ to change such that the probability of
the target label is increased and the probability of all other labels are
decreased. This method is notable for producing examples that have only been
changed in several dimensions, which can be hard for humans to detect. For a
full description of the attack, we refer the reader to \citet{jsma}.

\subsection{Defenses}

As baseline defenses, we consider defensive distillation and adversarial training.
To simplify comparison, we omit defenses \citep{xu2017feature,nayebi2017biologically} that are not fully
architecture-agnostic or which work by detecting and rejecting adversarial examples.

\subsubsection{Distillation}

Distillation, originally introduced by \citet{ba2014distillation},
was first examined as a potential
defense by \citet{distillation}. The main idea is that we train the model twice, initially using the one-hot ground truth labels but ultimately using the initial model's softmax probability
outputs, which contain additional information about the problem. Since the normal softmax
function tends to converge very quickly to one-hot-ness, we divide all of the logit network outputs
(which we will call $\hat{z}_k$ instead of the
probabilities $\hat{y}_k$) by a temperature $T$ (during training but not evaluation): \begin{equation}
  f_{T,\theta}(X_n)_k = \frac{e^{\hat{z}_k(X_n)/T}}{\sum_{i=1}^K e^{\hat{z}_i(X_n)/T}},
\end{equation}
where we use $f_{T,\theta}$ to denote a network ending in a softmax with temperature $T$.
Note that as $T$ approaches $\infty$, the predictions converge to
$\frac{1}{K}$. The full process can be expressed as \begin{equation}
\begin{split}
  \theta^0 & = \argmin_\theta \sum_{n=1}^N \sum_{k=1}^K -y_{nk} \log f_{T,\theta}(X_n)_k, \\
  \theta^* & = \argmin_\theta \sum_{n=1}^N \sum_{k=1}^K -f_{T,\theta^0}(X_n)_k \log f_{T,\theta}(X_n)_k.
\end{split}
\label{eq:distillation}
\end{equation}
Distillation is usually used to help small networks achieve the same accuracy as larger DNNs,
but in a defensive context, we use the same model twice.
It has been shown to be an effective defense against white-box FGSM attacks,
but \citet{carlini2016defensive} have shown that
it is not robust to all kinds of attacks. We will see that the precise way it
defends against certain attacks is qualitatively different than gradient regularization, and that it can actually make the models more vulnerable to attacks than an undefended model.

\subsubsection{Adversarial Training}

In adversarial training \citep{kurakin2016adversarial}, we increase robustness
by injecting adversarial examples into the training procedure. We follow the method implemented in
\citet{papernot2016cleverhans},
where we augment the network to run the FGSM on the training batches and
compute the model's loss function as the average of its loss on normal and
adversarial examples without allowing gradients to propogate so as to weaken the FGSM attack (which would also make the method second-order). We compute FGSM perturbations with respect to
predicted rather than true labels to prevent ``label leaking,'' where our model
learns to classify adversarial examples more accurately than regular examples.

\section{Gradient Regularization}

We defined our ``right for the right reasons'' objective in Chapter \ref{ch:1} using an L2 penalty on the gradient of the model's predictions across classes with respect to input features marked irrelevant by domain experts. We encoded their domain knowledge using an annotation matrix $A$. If we set $A=1$, however, and consider only the log-probabilities of the predicted classes, we recover what \citet{doublebackprop} introduced
as ``double backpropagation'', which trains neural networks by minimizing not just the ``energy'' of the network but the
rate of change of that energy with respect to the input features.
In their formulation the energy is a
quadratic loss, but we can reformulate it almost equivalently using the
cross-entropy: \begin{equation}
  \theta^*
      = \argmin_\theta \sum_{n=1}^N \sum_{k=1}^K -y_{nk} \log f_\theta(X_n)_k
     +  \lambda \sum_{d=1}^D \sum_{n=1}^N \left(\frac{\partial}{\partial x_d} \sum_{k=1}^K -y_{nk} \log f_\theta(X_n)_k\right)^2,
\label{eq:doubleback}
\end{equation}
whose objective we can write a bit more concisely as \[
  \argmin_\theta H(y, \hat{y}) + \lambda ||\nabla_x H(y, \hat{y})||_2^2,
\]
where $\lambda$ is again a hyperparameter specifying the penalty strength.
The intuitive objective of this function is to ensure that if any input changes slightly, the
divergence between the predictions and the labels will not change significantly (though including this term does not guarantee Lipschitz continuity everywhere). Double
backpropagation was mentioned as a potential adversarial defense in the same
paper which introduced defensive distillation \citep{distillation}, but at publish time, its effectiveness in this respect had not yet been analyzed in the
literature -- though \cite{gu2014towards} previously and \cite{hein2017formal,czarnecki2017sobolev} concurrently consider related objectives, and \cite{raghunathan2018certified} derive and minimze an upper bound on adversarial vulnerability based on the maximum gradient norm in a ball around each training input. These works also provide stronger theoretical explanations for why input gradient regularization is effective, though they do not analyze its relationship to model interpretability. In this work, we interpret gradient regularization as a quadratic penalty on our model's saliency map.

\section{Experiments}

\subsubsection{Datasets and Models}

We evaluated the robustness of distillation, adversarial training, and
gradient regularization to the FGSM, TGSM, and JSMA on MNIST \citep{mnist}, Street-View House Numbers (SVHN)
\citep{svhn}, and notMNIST \cite{not-mnist}.  On all datasets, we test a simple
convolutional neural network with 5x5x32 and 5x5x64 convolutional layers
followed by 2x2 max pooling and a 1024-unit fully connected layer, with
batch-normalization after all convolutions and both batch-normalization and
dropout on the fully-connected layer. All models were implemented in
Tensorflow and trained using Adam \citep{adam} with $\alpha=0.0002$ and $\epsilon=10^{-4}$ for 15000 minibatches of size of 256.
For SVHN, we prepare training and validation set as described in
\citet{sermanet2012convolutional}, converting the images to grayscale following
\citet{grundland2007decolorize} and applying both global and local contrast
normalization.

\subsubsection{Attacks and Defenses}

\begin{figure}[ht]
  \centering
  \includegraphics[width=0.67\textwidth]{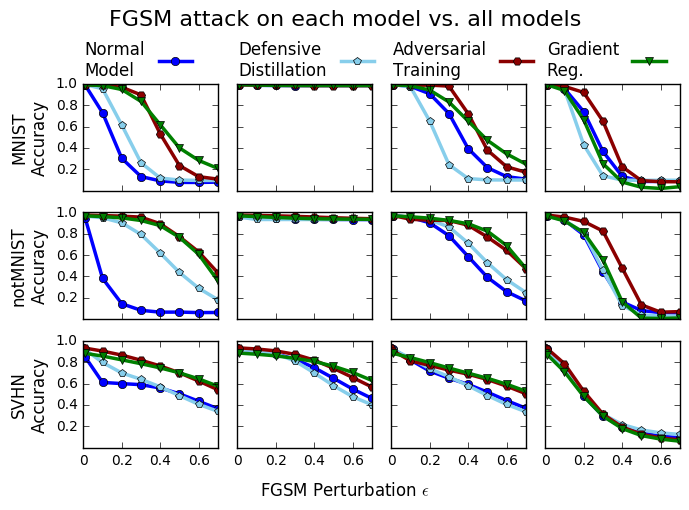}
  \caption{Accuracy of all CNNs on FGSM examples generated to fool undefended
    models, defensively distilled, adversarially trained, and
    gradient regularized models (from left to right) on MNIST, SVHN, and
    notMNIST (from top to bottom). Gradient-regularized models are the most
    resistant to other models' adversarial examples at high $\epsilon$, while all models are fooled
    by gradient-regularized model examples. On MNIST and notMNIST, distilled model examples are usually identical to non-adversarial examples (due to gradient underflow), so they fail to fool any of the other models.
  }
\label{fig:fgsm-comp}
\end{figure}

\begin{figure}[ht]
  \centering
  \includegraphics[width=0.67\textwidth]{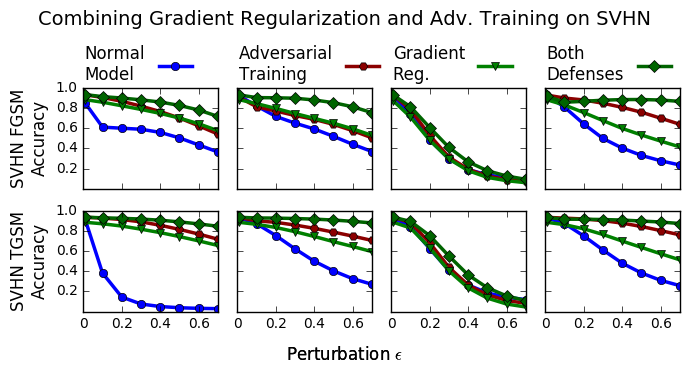}
  \caption{Applying both gradient regularization and adversarial training (``both defenses'') allows us to obtain maximal robustness to white-box and normal black-box attacks on SVHN (with a very slight label-leaking effect on the FGSM, perhaps due to the inclusion of the $\nabla_x H(y,\hat{y})$ term). However, no models are able to maintain robustness to black-box attacks using gradient regularization.}
\label{fig:both-defenses}
\end{figure}

For adversarial training and JSMA example generation, we used the Cleverhans adversarial example library
\citep{papernot2016cleverhans}. For distillation, we used a softmax temperature
of $T=50$, and for adversarial training, we trained with FGSM perturbations at
$\epsilon=0.3$, averaging normal and adversarial losses.
For gradient regularized models, we use double backpropagation, which provided the best robustness, and train over a spread of $\lambda$ values. We choose the $\lambda$ with the highest accuracy against validation black-box FGSM
examples but which is still at least 97\% as accurate on normal validation
examples (though accuracy on normal examples tended not to be significantly different). Code for all models and experiments has been open-sourced\footnote{\url{https://github.com/dtak/adversarial-robustness-public}}.

\subsubsection{Evaluation Metrics}

For the FGSM and TGSM, we test \textit{all} models
against adversarial examples generated for \textit{each} model and report accuracy.
Testing this way allows us to simultaneously measure white- and black-box robustness.

On the JSMA and iterated TGSM, we found that measuring accuracy was no longer
a good evaluation metric, since for our gradient-regularized models, the
generated adversarial examples often resembled their targets more than their
original labels. To investigate this, we performed a human subject experiment
to evaluate the legitimacy of adversarial example misclassifications.

\subsection{Accuracy Evaluations (FGSM and TGSM)}

\subsubsection{FGSM Robustness}

Figure \ref{fig:fgsm-comp} shows the results of our defenses' robustness to the
FGSM on MNIST, SVHN, and notMNIST for our CNN at a variety of perturbation strengths
$\epsilon$.
Consistently across datasets, we find that gradient-regularized models exhibit
strong robustness to black-box transferred FGSM attacks (examples produced by attacking other models). Although adversarial training sometimes performs slightly better at
$\epsilon \leq 0.3$, the value we used in training, gradient regularization generally surpasses it at higher $\epsilon$ (see the green curves in the leftmost plots).

The story with white-box attacks is more interesting. Gradient-regularized
models are generally more robust to than undefended models (visually, the green
curves in the rightmost plots fall more slowly than the blue curves in the
leftmost plots). However, accuracy still eventually falls for them, and it does
so faster than for adversarial training. Even though their robustness to
white-box attacks seems lower, though, the examples produced by those white-box
attacks actually fool \textit{all} other models equally well. This effect is
particularly pronounced on SVHN.  In this respect, gradient regularization may
hold promise not just as a defense but as an \textit{attack}, if examples
generated to fool them are inherently more transferable.

Models trained with defensive distillation in general perform no better and
often worse than undefended models. Remarkably, except on SVHN, attacks against
distilled models actually fail to fool all models. Closer inspection of
distilled model gradients and examples themselves reveals that this occurs
because distilled FGSM gradients vanish -- so the examples are not perturbed at
all. As soon as we obtain a nonzero perturbation from a different model,
distillation's appearance of robustness vanishes as well.

\begin{figure}[htb]
  \centering
  \includegraphics[width=\textwidth]{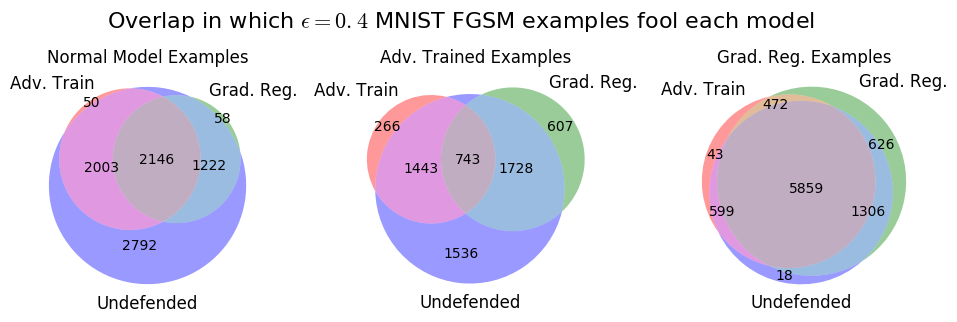}
  \caption{Venn diagrams showing overlap in which MNIST $\epsilon=0.4$ FGSM examples, generated for normal, adversarially trained, and gradient regularized models, fool all three. Undefended models tend to be fooled by examples from all models, while the sets of adversarially trained model FGSM examples that fool the two defended models are closer to disjoint. Gradient-regularized model FGSM examples fool all models. These results suggest that ensembling different forms of defense may be effective in defending against black box attacks (unless those black box attacks use a gradient-regularized proxy).}
  \label{fig:overlaps}
\end{figure}

\begin{figure}[htb]
  \centering
  \includegraphics[width=\textwidth]{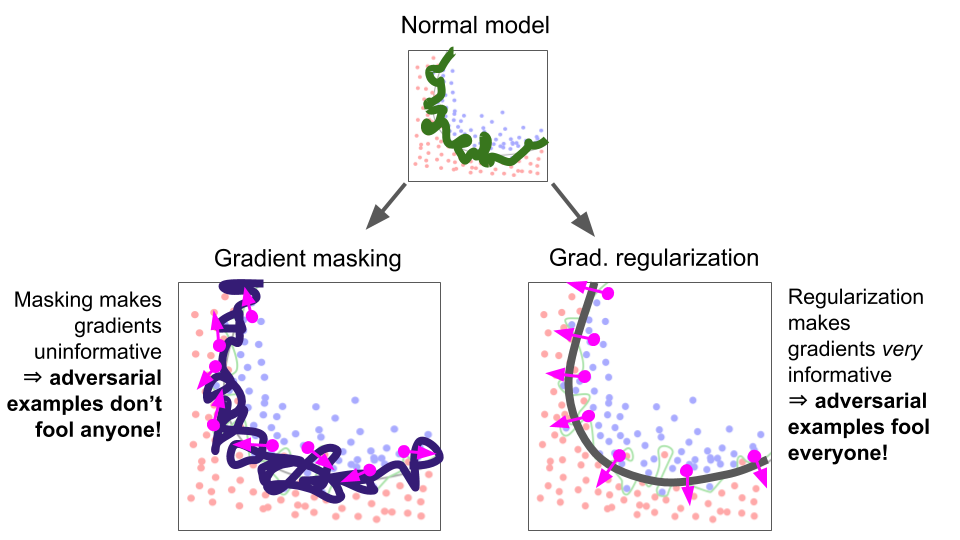}
  \caption{Conceptual illustration of the difference between gradient regularization and gradient masking. In (idealized) gradient masking, input gradients are completely uninformative, so following them doesn't affect either the masked model's predictions or those of any other model. In gradient regularization, gradients actually become \textit{more} informative, so following them will ultimately fool \textit{all} models. However, because gradients are also smaller, perturbations need to be larger to flip predictions. Unregularized, unmasked models are somewhere in between. We see quantitative support for this interpretation in Figure \ref{fig:overlaps}, as well as qualitative evidence in Figure \ref{fig:uncertainty-gradients}.}
  \label{fig:overlaps-intuition}
\end{figure}

Although adversarial training and gradient regularization seem comparable in
terms of accuracy, they work for different reasons and can be applied in
concert to increase robustness, which we show in Figure
\ref{fig:both-defenses}. In Figure \ref{fig:overlaps} we also show that, on
normal and adversarially trained black-box FGSM attacks, models trained with
these two defenses are fooled by different sets of adversarial examples. We provide intuition for why this might be the case in Figure \ref{fig:overlaps-intuition}.

\begin{figure}[htb]
  \centering
  \includegraphics[width=0.67\textwidth]{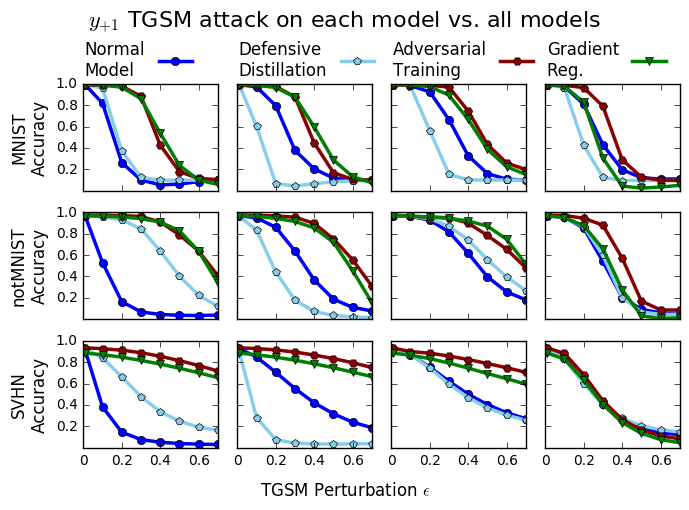}
  \caption{CNN accuracy on $y_{+1}$ TGSM examples generated to fool the four
  models on three datasets (see Figure \ref{fig:fgsm-comp} for more
explanation). Gradient-regularized models again exhibit robustness to
other models' adversarial examples. Distilled model adversarial perturbations fool other models again since their input gradients no longer underflow.}
\label{fig:tgsm-comp}
\end{figure}

\subsubsection{TGSM Robustness}

Against the TGSM attack (Figure \ref{fig:tgsm-comp}), defensively distilled
model gradients no longer vanish, and accordingly these models start to show
the same vulnerability to adversarial attacks as others.  Gradient-regularized
models still exhibit the same robustness even at large perturbations
$\epsilon$, and again, examples generated to fool them fool other models
equally well.

\begin{figure}[htb]
  \centering
  \includegraphics[width=0.49\textwidth]{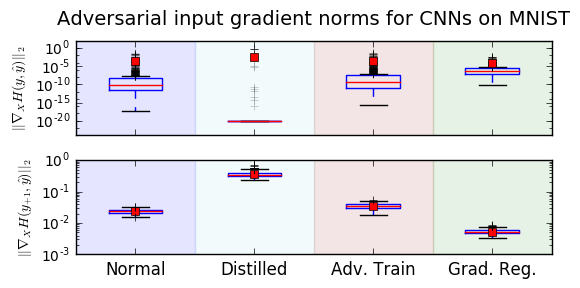}
  \includegraphics[width=0.49\textwidth]{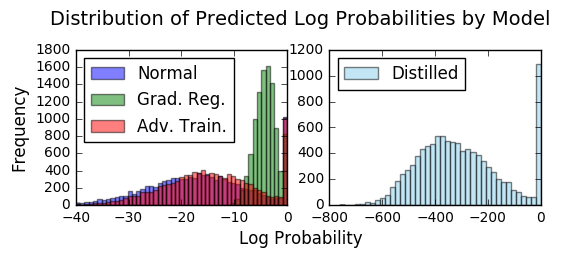}
  \caption{Distributions of (L2 norm) magnitudes of FGSM input gradients (top), TGSM input gradients (middle),
    and predicted log probabilities across all classes (bottom) for each defense. Note the logarithmic scales. Gradient-regularized models
    tend to assign non-predicted classes higher probabilities, and the L2 norms of the input gradients of their FGSM and TGSM
  loss function terms have similar orders of magnitude. Distilled models (evaluated at $T=0$) assign extremely small probabilities to all but the predicted class, and their TGSM gradients explode while their FGSM gradients vanish (we set a minimum value of $10^{-20}$ to prevent underflow). Normal and adversarially trained models lie somewhere in the middle.}
\label{fig:mnist-gradient-comparison}
\end{figure}

One way to better understand the differences between gradient-regularized, normal,
and distilled models is to examine the log probabilities they output and the norms of their loss function input gradients, whose
distributions we show in Figure \ref{fig:mnist-gradient-comparison} for MNIST. We can see that the different defenses have very different statistics.
Probabilities of non-predicted classes tend to be small but remain nonzero for gradient-regularized models,
while they vanish on defensively distilled models evaluated at $T=0$ (despite distillation's stated purpose of discouraging certainty). Perhaps because $\nabla \log p(x) = \frac{1}{p(x)}\nabla p(x)$, defensively distilled models' non-predicted log probability input gradients are the largest by many orders of magnitude, while gradient-regularized models' remain controlled, with much smaller means and variances. The other models lie between these two extremes.
While we do not have a strong theoretical argument about what input gradient magnitudes
\textit{should} be, we believe it makes intuitive sense that having less
variable, well-behaved, and non-vanishing input gradients should be associated
with robustness to attacks that consist of small perturbations in input space.

\begin{figure}[htb]
  \centering
  \includegraphics[width=0.49\textwidth]{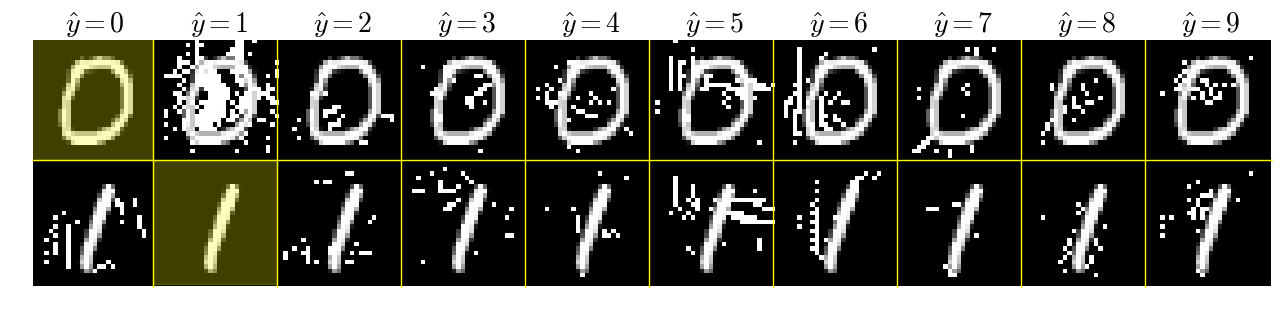}
  \includegraphics[width=0.49\textwidth]{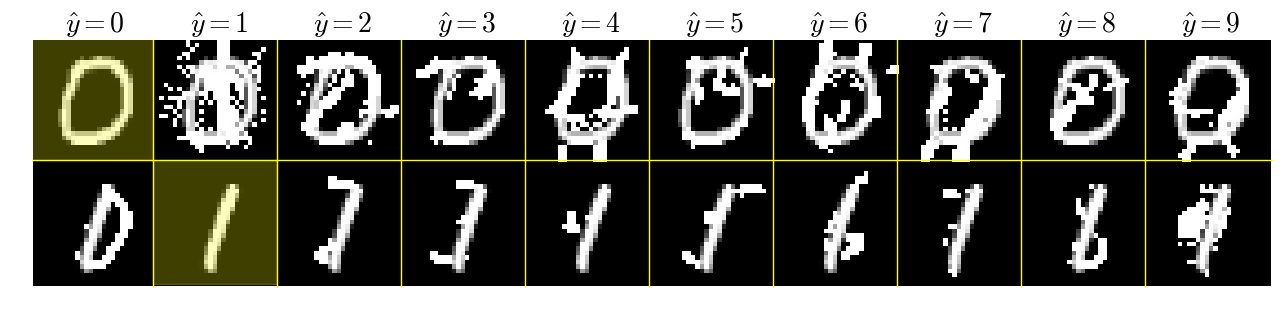}
  \caption{Results of applying the JSMA to MNIST \texttt{0} and \texttt{1}
  images with maximum distortion parameter $\gamma=0.25$ for a distilled model
(left) and a gradient-regularized model (right). Examples in each row start
out as the highlighted digit but are modified until the model predicts the
digit corresponding to their column or the maximum distortion is reached.}
  \label{fig:mnist-jsma-grid}
\end{figure}

\subsection{Human Subject Study (JSMA and Iterated TGSM)}

\subsubsection{Need for a Study}

Accuracy scores against the JSMA can be misleading, since without a
maximum distortion constraint it necessarily runs until the model predicts the
target.  Even with such a constraint, the perturbations it creates sometimes
alter the examples so much that they no longer resemble their original labels,
and in some cases bear a greater resemblance to their targets.  Figure
\ref{fig:mnist-jsma-grid} shows JSMA examples on MNIST for
gradient-regularized and distilled models which attempt to convert \texttt{0}s
and \texttt{1}s into every other digit.  Although all of the perturbations
``succeed'' in changing the model's prediction, in the
gradient-regularized case, many of the JSMA examples strongly resemble their
targets.

\begin{figure}[ht]
  \centering

  \includegraphics[width=0.32\textwidth]{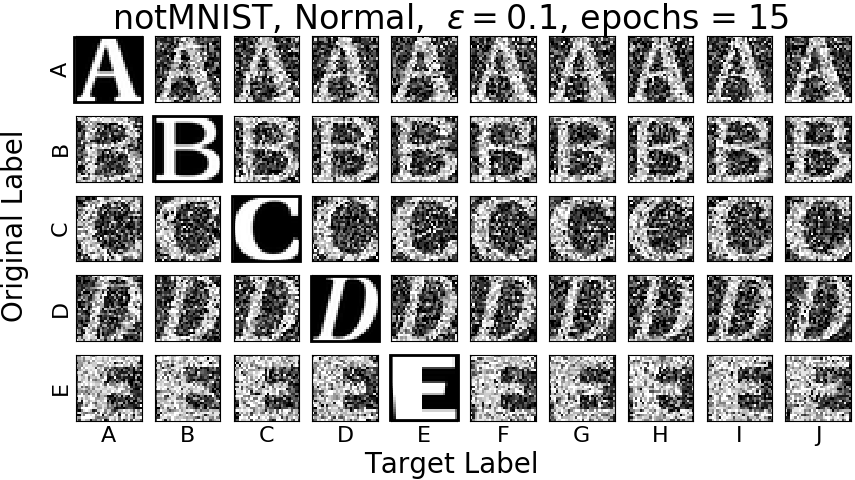}
  \includegraphics[width=0.32\textwidth]{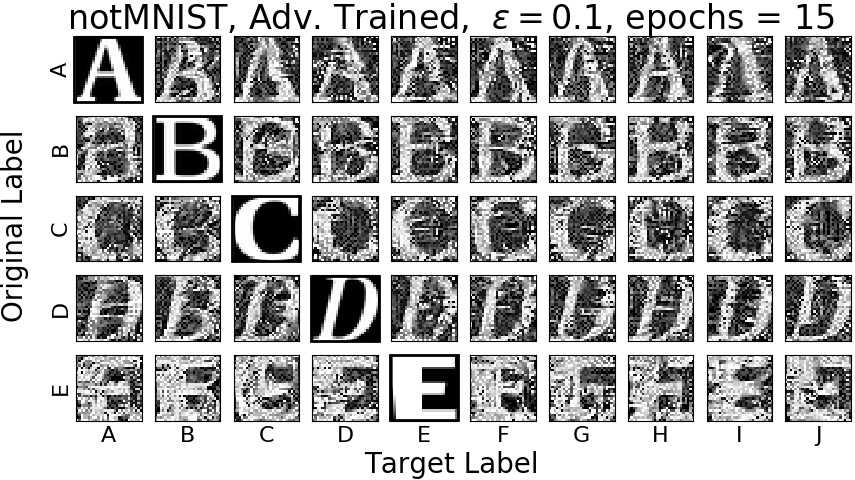}
  \includegraphics[width=0.32\textwidth]{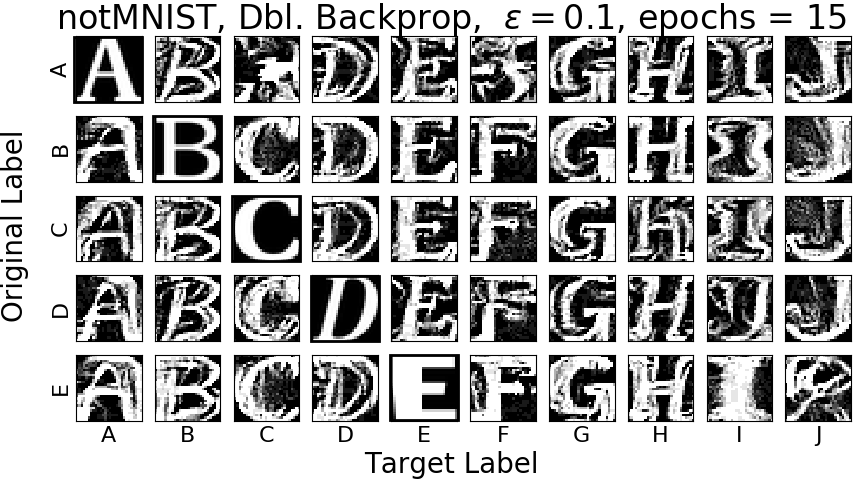}

  \includegraphics[width=0.32\textwidth]{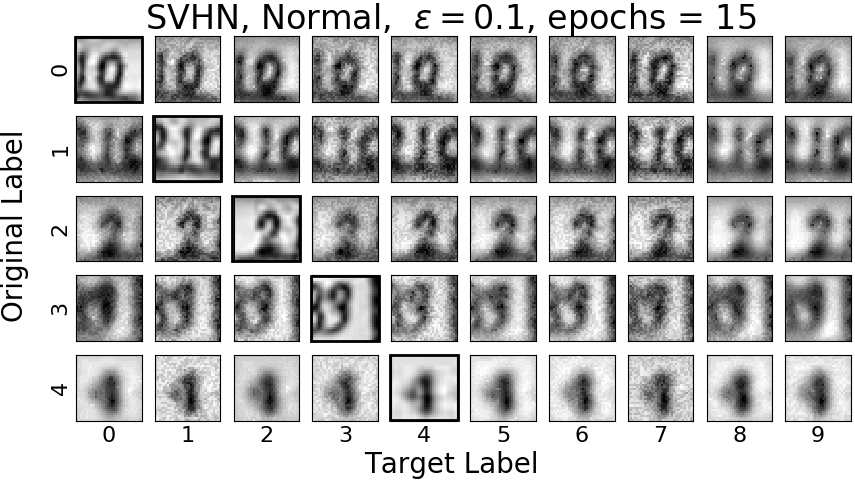}
  \includegraphics[width=0.32\textwidth]{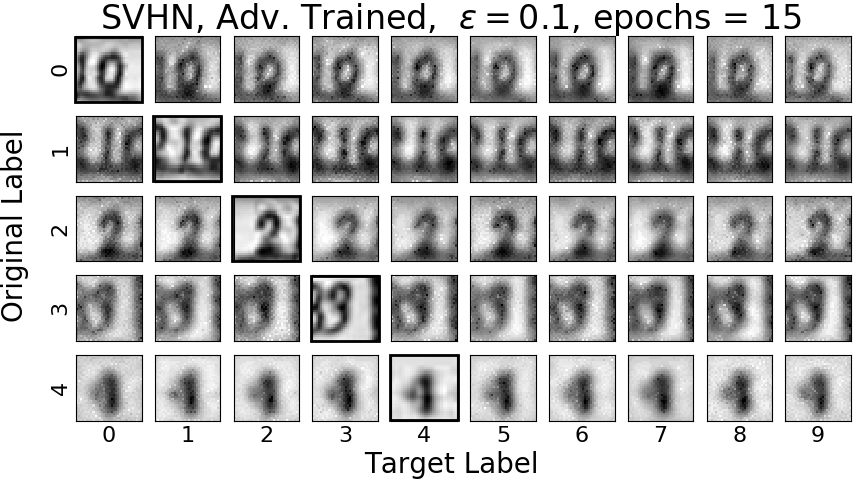}
  \includegraphics[width=0.32\textwidth]{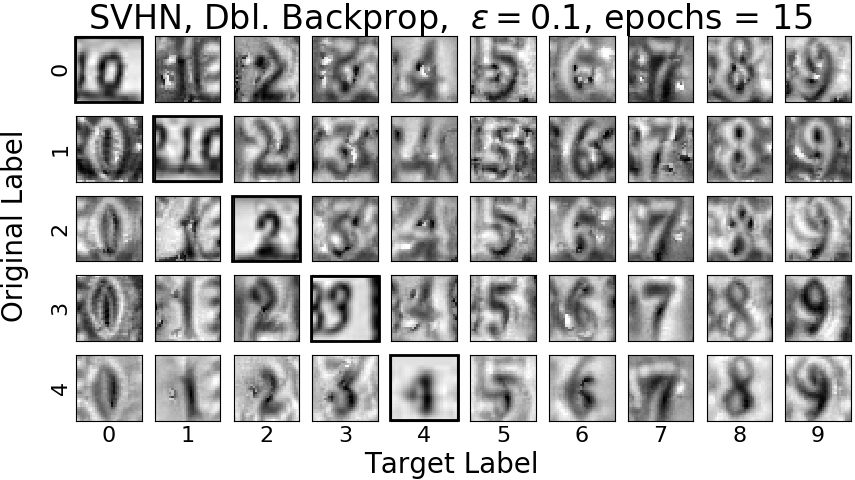}

  \caption{
    Partial confusion matrices showing results of applying the iterated TGSM for 15 iterations at $\epsilon=0.1$. Each row is generated from the same example but modified to make the model to predict every other class. TGSM examples generated for gradient-regularized models (right) resemble their targets more than their original labels and may provide insight into what the model has learned. Animated versions of these examples can be seen at \url{http://goo.gl/q8ZM1T}.
}
  \label{fig:tgsm-grids}
\end{figure}

The same issues occur for other attack methods, particularly the iterated TGSM,
for which we show confusion matrices for different models and datasets in Figure \ref{fig:tgsm-grids}.
For the gradient-regularized models, these psuedo-adversarial examples
quickly become almost prototypical examples of their targets, which is not
reflected in accuracies with respect to the original labels.

To test these intuitions more rigorously, we ran a small pilot study with 11
subjects to measure whether they found examples generated by these methods to
be more or less plausible instances of their targets.

\subsubsection{Study Protocol}

The pilot study consisted of a quantitative and qualitative portion. In the
quantitative portion, subjects were shown 30 images of MNIST JSMA or SVHN
iterated TGSM examples. Each of the 30 images corresponded to one original
digit (from 0 to 9) and one model (distilled, gradient-regularized, or
undefended). Note that for this experiment, we used $\nabla_x H(\frac{1}{K}, \hat{y})$ gradient regularization, ran the TGSM for just 10 steps, and trained models for 4 epochs at a learning rate of 0.001. This procedure was sufficient to produce examples with explanations similar to the longer training procedure used in our earlier experiments, and actually increased the robustness of the undefended models (adversarial accuracy tends to fall with training iteration).
Images were chosen uniformly at random from a
larger set of 45 examples that corresponded to the first 5 images of the
original digit in the test set transformed using the JSMA or iterated TGSM to
each of the other 9 digits (we ensured that all models misclassified all
examples as their target). Subjects were not given the original label, but were
asked to input what they considered the most and second-most plausible
predictions for the image that they thought a reasonable classifier would make
(entering N/A if they thought no label was a plausible choice).  In the
qualitative portion that came afterwards, users were shown three 10x10
confusion matrices for the different defenses on MNIST (Figure
\ref{fig:mnist-jsma-grid} shows the first two rows) and were asked to write
comments about the differences between the examples.  Afterwards, there was a
short group discussion. This study was performed in compliance with the
institution's IRB.

\begin{table}
  \begin{center}
  \begin{tabular}{|c|p{0.45in}|p{0.65in}|p{0.45in}|p{0.65in}|} \hline
 & \multicolumn{2}{c|}{MNIST (JSMA)} & \multicolumn{2}{|c|}{SVHN (TGSM)} \\
\hline
Model & human\newline fooled &   mistake\newline reasonable   & human\newline fooled & mistake\newline reasonable  \\
\hline
normal      & 2.0\%           & 26.0\%           & 40.0\%          & 63.3\%         \\
\hline
distilled   & 0.0\%           & 23.5\%           & 1.7\%           & 25.4\%          \\
\hline
grad. reg. & \textbf{16.4\%} & \textbf{41.8\%}  & \textbf{46.3\%} & \textbf{81.5\%}  \\
\hline
\end{tabular}
\end{center}
\caption{Quantitative feedback from the human subject experiment. ``human fooled'' columns record what percentage of examples were classified by humans as \textit{most} plausibly their adversarial targets, and ``mistake reasonable'' records how often humans either rated the target plausible or marked the image unrecognizable as any label (N/A).}
\label{table:human-study}
\end{table}

\subsubsection{Study Results}

Table \ref{table:human-study} shows quantitative results from the human subject
experiment. Overall, subjects found gradient-regularized model adversarial examples most
convincing. On SVHN and especially MNIST, humans were most likely to think that
gradient-regularized (rather than distilled or normal) adversarial examples were
best classified as their target rather than their original digit.
Additionally, when they did not consider the target the most plausible
label, they were most likely to consider gradient-regularized model mispredictions
``reasonable'' (which we define in Table \ref{table:human-study}), and more
likely to consider distilled model mispredictions unreasonable.
p-values for the differences between normal and gradient regularized unreasonable error rates were 0.07 for MNIST and 0.08 for SVHN.

In the qualitative portion of the study (comparing MNIST JSMA examples), \textit{all} of the written
responses described significant differences between the insensitive model's
JSMA examples and those of the other two methods. Many of the examples for the
gradient-regularized model were described as ``actually fairly convincing,'' and that the
normal and distilled models ``seem to be most easily fooled by adding spurious
noise.'' Few commentators indicated any differences between the normal and
distilled examples, with several saying that ``there doesn't seem to be [a]
stark difference'' or that they ``couldn't describe the difference'' between
them. In the group discussion one subject remarked on how the perturbations to
the gradient-regularized model felt ``more intentional'', and others commented on how
certain transitions between digits led to very plausible fakes while others
seemed inherently harder. Although the study was small, both its quantitative and qualitative
results support the claim that gradient regularization, at least for the two
CNNs on MNIST and SVHN, is a credible defense against the JSMA and the iterated
TGSM, and that distillation is not.

\subsection{Connections to Interpretability}

\begin{figure*}[htb]
  \centering
  \parbox{0.32\linewidth} {
    \includegraphics[width=\linewidth]{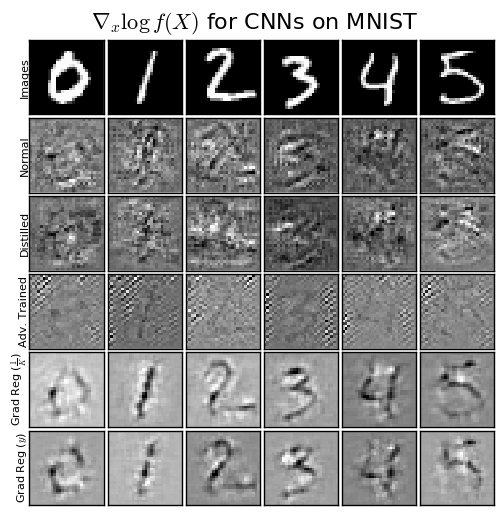}
  }
  \parbox{0.32\linewidth} {
    \includegraphics[width=\linewidth]{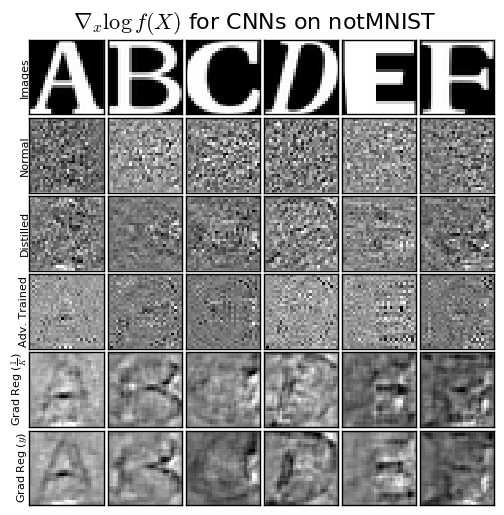}
  }
  \parbox{0.32\linewidth} {
    \includegraphics[width=\linewidth]{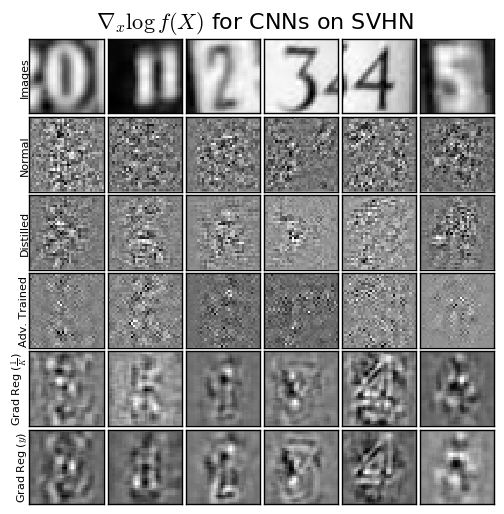}
  }
  \caption{Input gradients $\nabla_x H(\frac{1}{K},\hat{y})$
    that provide a local linear approximation of normal
    models (top), distilled models at $T=50$ (second from top), adversarially trained models (middle),
    and models trained with $\nabla_x H(\frac{1}{K},\hat{y})$ and $\nabla_x H(y,\hat{y})$
    gradient regularization (bottom two). Whitening black pixels or darkening
    white pixels makes the model more certain of its prediction. In general, regularized model
    gradients appear smoother and make more intuitive sense as local linear
    approximations.}
  \label{fig:uncertainty-gradients}
\end{figure*}
Finally, we present a qualitative evaluation suggesting a connection between
adversarial robustness and interpretability.  In the literature on
explanations, input gradients are frequently used as explanations
\citep{baehrens2010explain}, but sometimes they are noisy and not interpretable
on their own. In those cases, smoothing techniques have been developed
\citep{smoothgrad,deeplift,integrated-gradients} to generate more interpretable
explanations, but we have already argued that these techniques may obscure
information about the model's sensitivity to background features.

We hypothesized that if the models had more interpretable input gradients without the need for
smoothing, then perhaps their adversarial examples, which are generated
directly from their input gradients, would be more interpretable as well.  That is, the adversarial example would be more obviously transformative away from the original class label and towards
another. The results of the user study show that our gradient-regularized
models have this property; here we ask if the gradients are more interpretable as
explanations.

In Figure \ref{fig:uncertainty-gradients} we visualize input gradients across
models and datasets, and while we cannot make any quantitative claims, there
does appear to be a qualitative difference in the interpretability of the input
gradients between the gradient-regularized models (which were relatively robust
to adversarial examples) and the normal and distilled models (which were
vulnerable to them). Adversarially trained models seem to exhibit slightly more
interpretable gradients, but not nearly to the same degree as
gradient-regularized models.  When we repeatedly apply input gradient-based
perturbations using the iterated TGSM (Figure \ref{fig:tgsm-grids}), this
difference in interpretability between models is greatly magnified, and the
results for gradient-regularized models seem to provide insight into what the
model has learned. When gradients become interpretable, adversarial images
start resembling feature visualizations \cite{olah2017feature}; in other words,
they become explanations.

\section{Discussion}

In this chapter, we showed that:
\begin{itemize}
  \item
    Gradient regularization slightly outperforms adversarial training (the
    SOTA) as a defense against black-box transferred FGSM examples from
    undefended models.
  \item
    Gradient regularization significantly increases robustness to white-box
    attacks, though not quite as much as adversarial training.
  \item
    Adversarial examples generated to fool gradient-regularized models are more
    ``universal;'' they are more effective at fooling all models than examples
    from unregularized models.
  \item
    Adversarial examples generated to fool gradient-regularized models are more
    interpretable to humans, and examples generated from iterative attacks
    quickly come to legitimately resemble their targets. This is not true for
    distillation or adversarial training.
\end{itemize}
The conclusion that we would \textit{like} to reach is that gradient-regularized models
are right for better reasons. Although they are not completely robust to
attacks, their correct predictions and their mistakes are both easier to
understand. To fully test this assertion, we would need to run a larger and
more rigorous human subject evaluation that also tests adversarial training and
other attacks beyond the JSMA, FGSM, and TGSM.

Connecting what we have done back to the general idea of explanation
regularization, we saw in Equation \ref{eq:doubleback} that we could
interpret our defense as a quadratic penalty on our CNN's saliency map.  Imposing
this penalty had both quantitative and qualitative effects; our gradients became
smaller but also smoother with fewer high-frequency artifacts.  Since gradient
saliency maps are just normals to the model's decision surface, these changes
suggest a qualitative difference in the ``reasons'' behind our model's
predictions. Many techniques for generating smooth, simple saliency maps for
CNNs \textit{not} based on raw gradients have been shown to vary under
meaningless transformations of the model \cite{kindermans2017reliability} or,
more damningly, to remain invariant under extremely meaningful ones
\citep{adebayo2018local} -- which suggests that many of these methods either
oversimplify or aren't faithful to the models they are explaining.
Our approach in this chapter was, rather than simplifying our explanations of fixed models,
to optimize our \textit{models} to have simpler explanations.
Their increased robustness can be thought of as a useful side effect.

Although the problem of adversarial
robustness in deep neural networks is still very much an open one, these results may suggest a
deeper connection between it and interpretability. No matter what method proves most
effective in the general case, we suspect that any progress towards ensuring
either interpretability or adversarial robustness in deep neural networks will
likely represent progress towards both.